\documentclass[11pt]{amsart}
\usepackage[letterpaper,margin=1in]{geometry}
\usepackage{amsmath,amssymb,amsthm,mathtools,bm}
\usepackage{booktabs,enumitem,array}
\usepackage{algorithm}
\usepackage{algpseudocode}
\usepackage[numbers,sort&compress]{natbib}
\usepackage{microtype}
\usepackage{needspace}
\usepackage{graphicx}
\usepackage{float}
\usepackage[section]{placeins}
\makeatletter
\newcommand{\subsectionbarrier}{\begingroup\def\@fb@botlist{}\FloatBarrier\endgroup}
\makeatother
\usepackage{tikz}
\usetikzlibrary{arrows.meta,positioning,calc}
\graphicspath{{./}}
\usepackage[colorlinks=true,allcolors=blue,hypertexnames=false]{hyperref}
\hypersetup{pdftitle={The Price of Failure Risk: Clipped Barrier Policy Gradients with Convex Regularization},
 pdfauthor={Pekka Malo, Lauri Viitasaari, Antti Suominen, Eeva Vilkkumaa, Olli Tahvonen}}
\allowdisplaybreaks[2]
\newcommand{\E}{\mathbb E}
\newcommand{\Pprob}{\mathbb P}
\newcommand{\R}{\mathbb R}
\newcommand{\Ptwo}{\mathcal P_2(\R^d)}
\newcommand{\Hc}{\mathcal H}
\newcommand{\Jc}{\mathcal J}
\newcommand{\Wtwo}{W_2}
\newcommand{\KL}{D_{\mathrm{KL}}}
\newcommand{\ind}{\mathbf 1}

\newtheorem{theorem}{Theorem}

\newtheorem{proposition}[theorem]{Proposition}
\newtheorem{corollary}[theorem]{Corollary}
\theoremstyle{definition}
\newtheorem{assumption}{Assumption}
\newtheorem{definition}[theorem]{Definition}
\theoremstyle{remark}
\newtheorem{remark}[theorem]{Remark}
\newtheorem{example}{Example}

\title[Clipped barrier policy gradients with convex regularization]
{The Price of Failure Risk:\\ Clipped Barrier Policy Gradients\\ with Convex Regularization}
\author[P. Malo]{Pekka Malo$^1$}
\author[L. Viitasaari]{Lauri Viitasaari$^1$}
\author[A. Suominen]{Antti Suominen$^1$}
\author[E. Vilkkumaa]{Eeva Vilkkumaa$^1$}
\author[O. Tahvonen]{Olli Tahvonen$^2$}
\thanks{$^1$Department of Information and Service Management, Aalto University,
Espoo, Finland. $^2$Department of Forest Sciences, University of Helsinki, Helsinki, Finland.}
\newif\ifpreprintnote \preprintnotetrue
\ifpreprintnote
\thanks{A revised and corrected version of arXiv:2411.19193, which it replaces. Online Resource~1, with Appendices D--G, and Online Resource~2 are ancillary files of this arXiv record.}
\fi

\begin{document}
\begin{abstract}
We study policy learning when a decision maker limits the probability of exhausting a resource budget by a specified horizon. We add the remaining budget to the state, subtract an interior barrier from utility and clip the resulting reward below at $-m$. The expected discounted return equals stopped utility less the interior loss and a charge of $m/(1-\beta)$ per unit of geometric failure risk, where $\beta$ is the discount factor. An approximate maximizer is within its optimization error of the best utility net of interior loss among policies with no greater geometric risk. We bound its excess risk over a comparison policy in terms of interior loss, regularization cost and optimization error. The bound scales as $1/m$ when these terms are uniformly bounded. Risk can be assessed from a valid lower bound on value or from independent trajectories of the fixed policy. We prove exponential convergence of the exact population gradient flow when the parameter regularizer supplies sufficient curvature along generalized geodesics. The analysis covers relative entropy and convex potentials such as weight decay. For suitable potentials it also describes continuous-time training of a finite network. We bound the changes in value and risk when units are sampled independently from a softmax parameter law, and relate failure risk to the achieved action relative entropy to a reference policy. We illustrate the results with a grid-based wear-allowance benchmark, a learned lander and a numerically computed population gradient flow.
\end{abstract}
\maketitle
\section{Introduction}

Many operating decisions are constrained by a cumulative resource budget. A machine has a wear allowance until its next overhaul, a vehicle has a fuel tank, and a firm has an emissions allowance for the year. Actions generate utility while consuming the resource. Because consumption is uncertain, operating a machine more intensively can increase both its output and the risk of a breakdown before maintenance. We call the first budget exceedance a \emph{failure}. Where failure has a finite cost, such as downtime or a contractual penalty, the decision maker may accept a small probability of failure within a given horizon. Such a requirement is a chance constraint \citep{charnes1959}, which appears in finance as quantile hedging \citep{follmer1999} and in reinforcement learning (RL) as a bound on the risk of entering an error state \citep{geibel2005}. The question we study is how to learn a policy of high utility under this constraint and how to assess the failure risk of the policy that is deployed.

A common way to learn such a policy is to modify the Markov decision process presented to the learning algorithm. The remaining budgets are appended to the state \citep{saute2022,castellano2022}, and a penalty is charged at failure \citep{saute2022,massiani2022}. We subtract an interior barrier from utility, in the manner of the logarithmic barrier of interior-point methods \citep{fiacco1990}, and impose a finite lower bound on the combined one-step reward. We call this construction a \emph{clipped barrier}. Policy-gradient methods such as proximal policy optimization (PPO) \citep{ppo2017} or soft actor-critic (SAC) \citep{sac2018} can be applied without modification. In a simulator, where failure can be permitted, the construction is implemented as a layer between the learner and the environment. We call this layer a \emph{reward wrapper}.

The simplicity of this construction motivates studying the finite penalties used in practice. For almost-sure budgets, \citet{saute2022} relate the optimum under an infinite failure cost to the constrained optimum. For finite penalties they establish existence of optimal policies and convergence of optimal values as the penalty grows. We ask what a policy achieves at a given finite price, how far the policy returned by a learner can fall short, and what failure risk can be certified (Sect.~\ref{sec:related}). Existing mean-field analyses of the learning dynamics of softmax policies use relative entropy to a prior \citep{leahy2022} or no parameter regularization \citep{agazzi2020}. We show how to analyze convergence under a broader class of regularizers that includes convex potentials such as those used for weight decay.

Clipping is necessary because the unclipped interior barrier diverges at the boundary of the budget. If the action reference can exhaust a budget with positive probability, every policy whose action density is strictly positive relative to that reference has value $-\infty$ under the unclipped barrier. A finite penalty charged only at failure keeps returns finite and leaves utility unaltered before failure. We retain the interior barrier and clip the combined one-step reward, utility minus the barrier, from below at $-m$. The process enters an absorbing state at failure and incurs the charge $m$ in every period from failure onward. The clip bounds the loss on exploratory trajectories, and the interior barrier gives a direct incentive to keep a reserve before exhaustion.

We consider two kinds of regularization. In action regularization, a cost is assigned to the distribution of actions, for example its relative entropy to a reference distribution, as in the regularized Markov decision processes of \citet{geist2019}. In parameter regularization, a penalty can take the form of quadratic weight decay or \emph{anchoring}, a quadratic penalty on the distance from reference weights. The \emph{actor} is the parameterized policy. Its reference weights can come from a controller already in use. For the population analysis, we treat a network as a probability law over the parameters of its units, following the mean-field view of \citet{mei2018} and \citet{chizat2018}. In Sect.~\ref{sec:flow} we relate regularizers of this law to finite-network penalties.

\subsection{The problem}\label{sec:problem}

A controlled process has states $s_t$, actions $a_t$ chosen by a
policy $\pi$, a bounded utility $u(s_t,a_t)$ discounted by
$\beta\in(0,1)$, and resources with nonnegative costs and budgets. The
remaining reserves $z_t$ are carried in the state, and the failure time
$\tau$ is the first action index after which some reserve is negative.
For the admissible policy class $\Pi$, let
$V(\pi)=\E_\pi\sum_{t\ge0}\beta^tu(s_t,a_t)$.
The deployed policy must keep the probability of failure within its horizon $T$ below a tolerance $0<\alpha<1$, and the problem is
\begin{equation}\tag{$P_{T,\alpha}$}\label{prob:chance}
 \sup_{\pi\in\Pi} V(\pi)\qquad\text{subject to}\qquad
 p_T(\pi)=\Pprob_\pi(\tau\le T)\le\alpha .
\end{equation}
We specify the budget dynamics and the stochastic consumption in Sect.~\ref{sec:model}. For a task that ends after a fixed number of actions, we truncate the utility and the reward at that number (Sect.~\ref{sec:validation}).
The requirement is a chance constraint on a state-space event, the
reserve staying nonnegative through the horizon. Section~\ref{sec:when} compares it with almost-sure, expected-cost and tail-risk constraints.

\begin{figure}[tb]
 \centering
 \includegraphics[width=0.84\linewidth]{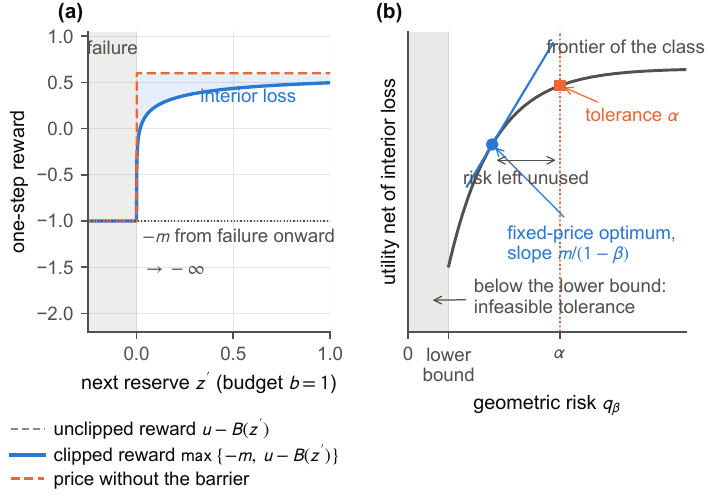}
 \caption{The logic of the clipped barrier. (a)~One-step reward against the next reserve $z'$ for an action of utility $u=0.6$, the barrier $B(z')=0.15\log(1+1/z')$ and $m=1$. The unclipped reward $u-B(z')$ tends to $-\infty$ at the boundary. Under the clipped barrier it is clipped at $-m$, a charge applied in every period from failure onward. Its interior loss (shaded) penalizes a small reserve before failure, which a price without the barrier does not. (b)~Schematic of the geometric risk against the utility net of interior loss for a policy class. The risk of every policy of the class is at least a positive number, a lower bound on the risk, and its frontier is the largest utility net of interior loss attainable at each risk. A price $m$ selects the point at which a line of slope $m/(1-\beta)$ supports the frontier, the tolerance $\alpha$ marks the feasible part of the frontier, and the priced optimum in this illustration leaves some risk unused. Below the lower bound the constraint is infeasible.}
 \label{fig:logic}
\end{figure}

In Sect.~\ref{sec:model} we define the objective that is trained. The one-step reward, utility less an interior barrier that is infinite at
the budget boundary, is clipped from below at a price $m$, and
failure is absorbing at that price. We show in Sect.~\ref{sec:risk} that
the resulting clipped value splits exactly into the utility
accumulated before failure, the interior loss subtracted by the barrier before failure, and the charge $m/(1-\beta)$ times the \emph{geometric risk}, the
probability of failure before an independent geometric time with
success probability $1-\beta$. The failure probability through action $T$ is at most $\beta^{-T}$ times the geometric risk. A finite price does not fix a risk level, so meeting a given tolerance requires this bound or validation on independent trajectories (Sect.~\ref{sec:validation}). We illustrate the construction in Fig.~\ref{fig:logic}, which distinguishes three decisions, namely keeping a reserve before failure, selecting an attained risk through the price, and meeting a given tolerance.

\subsection{Types of constraint in related formulations}
\label{sec:when}
Three common constraints differ in what they limit relative to the chance constraint in Eq.~\eqref{prob:chance}. Let the reserve follow $z_{t+1}=(z_t-c_t)/\gamma$ from $z_0=b$, with consumption $c_t\ge0$ and a cost discount $\gamma\in(0,1]$, so that the cumulative consumption $c_{\le T}=\sum_{t=0}^{T}\gamma^tc_t$ exceeds $b$ exactly when $\tau\le T$. For a limit $d$ and a tail mass $\varepsilon\in(0,1)$, the three constraints are
\begin{equation}\label{eq:neighbours}
 \Pprob_\pi(\tau\le T)=0,\qquad
 \E_\pi\sum_{t\ge0}\gamma^{t}c_t\le d,\qquad
 \mathrm{AV@R}_\varepsilon(c_{\le T})\le d,
\end{equation}
where $\mathrm{AV@R}_\varepsilon(Z)=\inf_{r\in\R}\{r+\varepsilon^{-1}\E[Z-r]^+\}$ is the conditional value-at-risk (CVaR), the mean of $Z$ over its worst $\varepsilon$-fraction \citep[(6.22)]{shapiro2009lectures}.

The first is the almost-sure constraint of safe sets, viability kernels and almost-sure budgets \citep{ames2019,cheng2019,massiani2022,saute2022,simmer2022}. It is the case $\alpha=0$. Bounded-score policies in Sect.~\ref{sec:model} cannot meet it when their action reference has positive probability of failure by $T$. The second is the expected-cost constraint of constrained Markov decision processes (constrained MDPs) \citep{altman1999,achiam2017}. For a mean-consumption limit $d$, Markov's inequality gives a bound $d/b$ on the probability of exceeding budget $b$. The expected indicator $\E_\pi\ind_{\{\tau\le T\}}$ equals $p_T(\pi)$ and gives \eqref{prob:chance} with limit $\alpha$. The discounted sum of the absorbing failure indicator has expectation $q_\beta(\pi)/(1-\beta)$, the cost that enters Sect.~\ref{sec:fixed-price}. The third constrains CVaR, a risk measure of the cumulative cost. Policy-gradient algorithms for this formulation include those of \citet{chow2018}. With the budget as the limit and the tail mass equal to the tolerance, it is a conservative approximation of \eqref{prob:chance} \citep[(6.24)]{shapiro2009lectures} that penalizes the overshoot over the budget. Almost-sure feasibility implies the chance constraint at every positive tolerance, and the CVaR constraint implies it at the matching tolerance, whereas the chance constraint restricts neither the mean nor the size of the consumption. Appendix~E.1 of Online Resource~1 relates the price to the multipliers, thresholds and tail masses of these formulations.

The deployment problem \eqref{prob:chance} uses continuing utility $V$ and the finite-horizon probability $p_T$. The clipped objective instead prices geometric risk against stopped utility net of interior loss. Up to an additive constant, the clipped objective is the Lagrangian of the geometric-risk problem at multiplier $m/(1-\beta)$. The geometric weight on a failure decreases exponentially with its time (Sect.~\ref{sec:risk}). The attained risk depends on the policy class and the policy returned at that price. Section~\ref{sec:prefer} states when another approach is preferable, for example when no failure is permissible or when the model is known and small.

\subsection{Contributions}
We establish what a finite clipped barrier provides when it is applied to a learner already in use, and what
the learner and the validation must still supply. Our main theoretical result is the population convergence theorem in contribution~(2).

(1)~We identify the clip level as a price of failure risk. The clipped value decomposes exactly into the utility earned before failure, the cost of keeping a reserve before failure and the discounted charge for failure risk. An approximate maximizer of the clipped value maximizes utility net of the interior loss, up to its optimization error, among policies with no greater geometric risk. For an exact maximizer, the discounted price supports the utility--risk frontier at the attained risk (Proposition~\ref{prop:fixed-price} and Eq.~\eqref{eq:price-supports-frontier}). A valid lower bound on the clipped value gives a risk certificate for any candidate policy (Proposition~\ref{prop:risk-value}). In a comparison with any represented policy of finite regularization cost, the interior loss, the regularization cost, the optimization error and the signed risk difference appear separately (Theorem~\ref{thm:matched-risk}). Since early failures are weighted exponentially more heavily than late ones, a bound on the geometric risk also bounds from below the average failure time of the earliest trajectories (Remark~\ref{rem:failure-time}). Optimality at the attained risk is a property of Lagrangian relaxation with a fixed multiplier \citep{altman1999}. The exact decomposition identifies the separate effects of keeping a reserve, pricing failure and regularizing the policy.

(2)~We prove a convergence theorem for the population policy-gradient flow under general convex parameter regularization (Theorem~\ref{thm:flow}), together with the separation of the roles of the regularizers that it makes possible. The theorem allows every parameter regularizer that is convex along generalized geodesics \citep{ags2005}, whereas \citet{leahy2022} treat relative entropy to a Gibbs prior. One argument therefore covers relative entropy, its displacement-convex extensions and convex potentials such as weight decay (Proposition~\ref{prop:generators}). Along interpolations of the units, the negative clipped value and the action relative entropy fall short of convexity by a deficit that grows linearly with the price, while the parameter regularizer supplies curvature. The condition of the theorem, which we call a \emph{curvature budget}, is that the supplied curvature exceeds the deficit, and under it the gradient flow converges exponentially to its unique minimizer (Corollary~\ref{thm:combined} bounds the optimization error and the risk along the gradient flow). Under a common smooth potential with Lipschitz gradient, width-scaled continuous-time training follows the population gradient flow from the network's empirical law and inherits the rate when $\Lambda_m>0$ (Proposition~\ref{prop:finite-training}), as \citet{chizat2018} observe for particle gradient flows.

The roles of the regularizers are visible in the particle system associated with the gradient flow (Sect.~\ref{subsec:prior}). Relative entropy gives rise to noise in each unit, which can move units out of regions where the value gradient vanishes. A quadratic potential pulls the units toward its center. Relative entropy ties the noise and the prior-induced drift to one strength, while the general class allows separate weighting. An integrand of faster growth limits how far the law of the units departs from the prior without changing the guaranteed rate or threshold. In the one-dimensional analytical example, unregularized transport leaves units on a plateau of the feature and the simulated units leave the plateau under relative entropy. The computed minimizers depend on the prior and have higher clipped value at smaller strengths in this example. On the action side, the geometric failure risk of a policy with action regularization stays within $\sqrt{\mathcal A/2}$ of the risk of a trusted reference, where $\mathcal A$ is the achieved action relative entropy to that reference (Proposition~\ref{prop:action-risk-bridge}). The sufficient curvature threshold grows with the price. We compare that threshold with the strengths used in the experiments.

A larger parameter weight can improve the guaranteed convergence rate while shifting the minimizer of the training objective. This trade-off motivates reducing the strength during training. With the price and action weight fixed, Proposition~\ref{prop:decreasing-strength} shows that if the parameter strength and the optimization gap both tend to zero, the objective values of the corresponding near-minimizers with the parameter term removed approach the infimum of that objective. Along the exact flow with stepwise reductions, the objective at the current strength is nonincreasing. The rate bound applies within stages above the curvature threshold. In our scheduling comparisons, the gain in the analytical example depends on the prior, and in the wear study the unanchored learner reaches the comparison utility sooner (Sect.~\ref{sec:schedules}).

(3)~We provide verifiable evidence for the deployed policy. We assess learned policies by independent validation, through a binomial test or a bound on the risk from the estimated value with Hoeffding's inequality \citep{hoeffding1963}. Both procedures use fresh trajectories of the fixed policy and apply independently of the population training guarantee (Sect.~\ref{sec:validation}). For softmax policies, we bound the change in value and risk when units are sampled independently from a parameter law to construct one network (Theorem~\ref{thm:finite-deployment}). We examine the relation between price and risk, compare fixed and adaptive multipliers, and quantify the effects of the barrier and the regularizers on a grid-based wear-allowance benchmark, a learned lander and the analytical example. Our companion toolbox implements the reward wrapper and the certificates around standard learners.

\subsection{Relation to existing work}\label{sec:related}
In control, a constraint on the state that has to hold with a prescribed confidence defines a problem of \emph{stochastic viability} \citep{doyen2010}, whose viable states and feedbacks are characterized by dynamic programming. In this paper the state includes the reserves, and clipping keeps the return finite for every policy even when failure has positive probability.

Constrained policy optimization (CPO) is a trust-region algorithm for expected-cost constrained MDPs \citep{achiam2017}. It handles the cost limit through a local constraint at each policy update. For the exact trust-region update, \citet{achiam2017} bound constraint violation up to a term of order equal to the square root of the trust-region radius. Chance and CVaR constraints on cumulative costs can also be treated by updating a Lagrange multiplier during training \citep{chow2018}. In our priced objective, $m/(1-\beta)$ is a fixed multiplier on geometric risk. The expected-cost problem solved by CPO has this objective as its fixed-multiplier Lagrangian when the utility, cost and policy class agree (Sect.~\ref{sec:prefer}). The risk bounds in Sect.~\ref{sec:risk} apply to the policy returned by a learning algorithm when the stated hypotheses hold.

For the almost-sure constraint, Saut\'e RL \citep{saute2022,simmer2022} uses the same augmentation by the budget, with a finite failure penalty as a proxy for an infinite one and without the barrier. Its guarantees concern the infinite penalty \citep[Theorem~3]{saute2022}, and for the finite penalty it gives the existence of an optimal policy and the convergence of the optimal values as the penalty grows \citep[Equation~(8), Theorem~2]{saute2022}. For the clipped barrier with a finite price $m$, we instead certify the risk of the policy actually returned. Safe sets, viability kernels and safe value functions \citep{ames2019,cheng2019,massiani2022,castellano2022} require a model, a known safe set or a learned feasibility function.

Action regularization in constrained RL \citep{ying2022} and joint action/parameter-law entropy in mean-field RL \citep{leahy2022} predate this study. Beyond extending the parameter regularizer of \citet[Theorem~2.12]{leahy2022} to every regularizer convex along generalized geodesics, we make explicit how the required regularization grows with the price $m$ and charge the regularization cost to the comparison law. Under relative entropy, \citet{lascu2025} obtain global convergence for a best-response gradient flow above a regularization-strength threshold.

We develop the model, the finite-price results, population convergence and deployment in Sects.~\ref{sec:model}--\ref{sec:finite-deployment}.
In Sect.~\ref{sec:numerics} we compare the bounds with computed values and report the learning results, and in Sect.~\ref{sec:guidance} we draw the practical implications.
We prove the central results in Appendices A--C. Online Resource~1 gives the test problems, experimental settings and further results. It also treats related formulations, proofs for the analytical example, examples of the risk bounds, a workflow and the toolbox.

\section{Budgets and the clipped-barrier mechanism}\label{sec:model}

The remaining reserves are part of the observed state. In this section
we state the model sketched in Sect.~\ref{sec:problem} precisely and
define the clipped reward and the represented policies. Their consequences are
then illustrated by Example~\ref{ex:running} in Sect.~\ref{sec:risk}.

\subsection{Model and deployment requirement}

Let $S,A$ be Polish spaces, $p_0$ the initial-state law, and
$Q(ds',dc\mid s,a)$ a Borel environmental kernel on
$S\times[0,\infty)^K$, where $K$ is the number of resources. After action $a_t$, the next physical state and a finite
nonnegative resource consumption $c$ are drawn from the kernel.
Consumption may contain an unobserved shock. We assume no knowledge of that shock before choosing $a_t$. Deterministic Borel costs
$c_k=g_k(s_t,a_t)$ are a special case, with $p(ds'\mid s,a)$ the
marginal transition.
The nonempty admissible action sets $A(s)$ have a Borel graph.
Assume $\Pi$ is a nonempty class of admissible Borel policy kernels
that choose actions based on the observed history.

For budgets $b_k>0$ and cost discounts $\gamma_k\in(0,1]$, the
augmented state is $y_t=(s_t,z_t)$ and
\begin{equation}\label{eq:budget}
 z_{0,k}=b_k,\qquad z_{t+1,k}=\frac{z_{t,k}-c_k}{\gamma_k},
\end{equation}
with $c$ the consumption drawn after $(s_t,a_t)$. The kernel $Q$ and
this recursion induce the augmented transition kernel
$\mathsf P(dy'\mid y,a)$.
Utility $u$ is Borel with $-u_-\le u\le\bar u$, $u_-,\bar u\ge0$.
Its discount is $\beta\in(0,1)$. Expectations include initial,
environmental and action randomness.
Failure occurs at the action index
$\tau=\inf\{t\ge0:\min_k z_{t+1,k}<0\}$, with $\inf\varnothing=\infty$.
Budget equality is not failure. Since costs are nonnegative, a
negative reserve stays negative. The finite-horizon and geometric
risks are
\[
 p_T(\pi)=\Pprob_\pi(\tau\le T),\qquad
 q_\beta(\pi)=\E_\pi[\beta^\tau\ind_{\{\tau<\infty\}}],
\]
where $T$ is the last action index, so $T+1$ actions are subject to the requirement.
The deployment problem is \eqref{prob:chance}, stated in the introduction.
The continuing utility $V$ is the deployment objective. In the decomposition of the clipped
return we use the stopped utility $U(\pi)=\E_\pi\sum_{t<\tau}\beta^tu(s_t,a_t)$, the utility before failure. We compare the two in Sect.~\ref{sec:performance},
with an explicit correction for the utility after failure.
The eventual requirement $\Pprob_\pi(\tau<\infty)\le\alpha$ is
different. A hard almost-sure budget is its special case $\alpha=0$.

\subsection{Clipping the combined reward}

Let $B:\R^K\to[0,\infty]$ be Borel, finite on $(0,\infty)^K$ and
infinite elsewhere, for example $B(z)=\theta_B\sum_k\log(1+b_k/z_k)$
on the positive orthant, with $\theta_B>0$. The \emph{clipped barrier reward}
is the one-step combined reward clipped from below at $-m$, with
the barrier evaluated at the updated reserve. The \emph{clipped value} is
its discounted expectation,
\begin{equation}\label{eq:clippedreward}
 r_m(y_t,a_t,z_{t+1})=
 \max\{-m,u(s_t,a_t)-B(z_{t+1})\},\qquad m>0,
\end{equation}
\begin{equation}\label{eq:Vm}
 V_m(\pi)=\E_\pi\sum_{t\ge0}\beta^t r_m(y_t,a_t,z_{t+1}),
 \qquad -m\le r_m\le\bar u .
\end{equation}
The interior barrier distinguishes small positive reserves from ample ones. At zero reserve the reward is clipped to $-m$, and the resulting loss enters the interior term because failure requires a strictly negative reserve.

Under a \emph{failure-only price}, the reward is $u(s_t,a_t)$ whenever every updated reserve is nonnegative, and $-m$ whenever any updated reserve is negative. Thus utility is paid at zero reserve and the charge begins at the first budget violation. If $m\ge u_-$, this reward can also be written as Eq.~\eqref{eq:clippedreward} with $B=0$ on the nonnegative orthant and $B=+\infty$ elsewhere.

From failure onward, every reward is $-m$, so $m$ is a loss in each period after failure, not a one-off charge at failure. Its discounted total from the failure time is $m\beta^\tau/(1-\beta)$, which equals a one-off charge of $m/(1-\beta)$ at $\tau$. A failure that costs a lump sum $F$ once is therefore represented by $m=(1-\beta)F$. We show in Sect.~\ref{sec:risk} that $m$ is a price of failure risk, with $m/(1-\beta)$ the cost of a unit of geometric risk. Below we call $m$ the \emph{price}. In a simulator, one may therefore terminate a failed trajectory and assign this charge at its failing action, and the value is then implemented exactly. For a task that physically ends after a fixed number of actions, $-m$ is charged only until its last action (Sect.~\ref{sec:validation}).

For the policy-gradient analysis, we integrate out the unobserved consumption draw in the conditional reward $\bar r_m(y,a)=\int r_m(y,a,z')\,\mathsf P(dy'\mid y,a)$. The policy still chooses its action from $y$ alone. The conditional reward is Borel with $-m\le\bar r_m\le\bar u$. We use the realized reward in the pathwise identities and the validation, and $\bar r_m$ in policy-averaged operators. In path formulas we abbreviate $r_m(y_t,a_t,z_{t+1})$ as $r_m(y_t,a_t)$.

\subsection{The represented class}

The certificates apply to any admissible Borel action kernel and do
not depend on a particular actor family. For population learning, we write
$\pi_\mu$ when the kernel is indexed by a law $\mu$ of network units.
We use the following fixed-reference softmax as the representative
construction for the explicit constants, the prior comparisons and the
sampling result. Other kernels enter the convergence theorem through
its derivative assumption, which must be verified for each actor map.

Let $\rho(da\mid y)$, the \emph{action reference}, be a fixed Borel
probability kernel supported on $A(s)$, and let $\psi(y,a,x)$ be
jointly measurable, with $|\psi|\le L_0$ for $x\in\R^d$. Let $\Ptwo$
denote the probability laws on $\R^d$ with finite second moment. For
$\mu\in\Ptwo$, set
\begin{equation}\label{eq:softmax}
 f_\mu(y,a)=\int\psi(y,a,x)\,\mu(dx),\qquad
 \pi_\mu(da\mid y)=
 \frac{\exp(f_\mu(y,a))\,\rho(da\mid y)}
      {\int\exp(f_\mu(y,a'))\,\rho(da'\mid y)} .
\end{equation}
The parameter law $\mu$ describes the units of one averaged-score
network. Its finite-width counterpart is constructed in
Sect.~\ref{sec:finite-deployment}. We assess the stochastic policy itself.
We write $V_m(\mu)$ or $q_\beta(\mu)$ for evaluation at $\pi_\mu$.
The density satisfies $e^{-2L_0}\le d\pi_\mu/d\rho\le e^{2L_0}$. By multiplying the density bounds along a path, we find that each exhaustion event by time $t$ has probability at least $e^{-2L_0(t+1)}$ times its probability under $\rho$. If the action reference can exhaust a budget with positive probability, so can every represented policy. With an unclipped barrier, the return is then $-\infty$ on an event of positive probability. Since the rest of the return is at most $\bar u/(1-\beta)$, every represented policy has the value $-\infty$. A lower bound $p_T(\pi_\mu)\ge e^{-2L_0(T+1)}p_T(\rho)$ on the risk follows from the same bounds, and no tolerance below it can be met. The objective is finite after clipping.

\section{Risk certificates and performance comparisons}\label{sec:risk}

The price enters the reward. We relate it to the risk and utility of the policy the learner returns. Proposition~\ref{prop:risk-value} bounds its risk from a valid lower bound on value. For the performance comparisons we also need an objective comparison or an optimization gap, written $\delta$ below. When we compare represented laws, we account separately for the cost of regularization. In Sect.~\ref{sec:flow} we bound the optimization gap of the population gradient flow under positive net curvature.

\subsection{The population objective and its risk interpretation}\label{sec:population-objective}

Training maximizes the clipped value while two regularization costs keep the learned policy near trusted references, one on the distribution of actions and one on the parameters of the network. A law $\mu\in\Ptwo$ on the parameters of the units defines the policy $\pi_\mu$ of Eq.~\eqref{eq:softmax}. For $m,\kappa>0$ and $\eta\ge0$, the regularization cost and the regularized objective are
\begin{equation}\label{eq:training-objective}
 \mathcal R(\mu)=\eta\mathcal A(\mu)+\kappa\Hc(\mu),
 \qquad \Jc_m(\mu)=-V_m(\mu)+\mathcal R(\mu),
 \quad\mu\in\mathcal M,
\end{equation}
where $\mathcal A$ is the action relative entropy and $\Hc$ the parameter regularizer defined below, with domain $\mathcal M=\{\mu:\Hc(\mu)<\infty\}$. We state the risk bounds in terms of the clipped value $V_m$ itself, not $\Jc_m$.

The action term measures how far the policy departs from the action reference $\rho$. With $k_\mu(y)=\KL(\pi_\mu(\cdot\mid y)\|\rho(\cdot\mid y))$, the \emph{action relative entropy}, stopped at failure, is
\begin{equation}\label{eq:action-regularizer-dynamic}
 \mathcal A(\mu)
 =\E_{\pi_\mu}\sum_{t=0}^{\tau}\beta^t k_\mu(y_t),
 \qquad 0\le\mathcal A(\mu)\le\frac{K_0}{1-\beta}.
\end{equation}
The sum includes the failing action, whose current budget is nonnegative. In the bound we assume $k_\mu(y)\le K_0$ uniformly, and for the softmax \eqref{eq:softmax} one can take $K_0=2L_0$. Without action regularization ($\eta=0$) the action term is absent, and the finiteness of $\mathcal A$ is required only for the separate behavioral comparisons that use it.

The parameter term measures how far the law of the units departs from a prior law or from reference parameters. The \emph{prior} $\Gamma\in\Ptwo$ is a reference law of unit parameters chosen before training, for instance a Gaussian centered at zero or at the parameters of a network already in use, the \emph{incumbent}, and its policy $\pi_\Gamma$ is the \emph{prior policy}. A \emph{parameter regularizer} is a functional $\Hc:\Ptwo\to[0,\infty]$ with nonempty domain. A \emph{divergence} is zero at the prior and grows with the distance from it, and a \emph{potential} is zero when every unit is at a reference and grows with the distance of the units from it. Parameter regularization is removed by setting $\Hc\equiv0$.

\begin{example}[Parameter regularizers used in practice]\label{ex:regularizers}
The relative entropy to the prior, $\KL(\mu\|\Gamma)$, is a divergence. Quadratic weight decay, $\int\frac12|x|^2\,d\mu$, and anchoring to reference parameters $x_0$, $\int\frac12|x-x_0|^2\,d\mu$, are potentials. On the empirical law $\mu_n=n^{-1}\sum_{i=1}^n\delta_{x_i}$ of a finite network the two potentials are the usual penalties $(2n)^{-1}\sum_i|x_i|^2$ and $(2n)^{-1}\sum_i|x_i-x_0|^2$, whereas the relative entropy to a continuous prior is infinite on $\mu_n$ (Sect.~\ref{subsec:prior}).
\end{example}

For a
bounded reward approximation $\|\widetilde r_m-r_m\|_\infty\le\varepsilon_r$,
put $\widetilde\Jc_m=-\widetilde V_m+\mathcal R$ with the regularizers
unchanged. Then
\begin{equation}\label{eq:direct-certificate-step0}
 |\widetilde V_m(\mu)-V_m(\mu)|\le\frac{\varepsilon_r}{1-\beta},
\end{equation}
and $\varepsilon_r=0$ recovers the exact objective. Let $\widehat\mu\in\mathcal M$ be a law with
\begin{equation}\label{eq:risk-inexact-optimum}
 \widetilde\Jc_m(\widehat\mu)
 \le\inf_{\mu\in\mathcal M}\widetilde\Jc_m(\mu)+\delta,
 \qquad\delta\ge0 .
\end{equation}
A \emph{comparison law} is a represented law $\bar\mu\in\mathcal M$
used to assess the learned policy, not necessarily an input to the
algorithm. The comparison results below hold for every comparison law simultaneously under Eq.~\eqref{eq:risk-inexact-optimum}, and for one specified
comparison law when Eq.~\eqref{eq:risk-inexact-optimum} is replaced by
$\widetilde\Jc_m(\widehat\mu)\le\widetilde\Jc_m(\bar\mu)+\delta$.

\begin{proposition}[Value and objective certificates]\label{prop:risk-value}
For every admissible policy $\pi$ and integer $T\ge0$,
\begin{equation}\label{eq:risk-horizon-bound}
 q_\beta(\pi)\le\frac{\bar u-(1-\beta)V_m(\pi)}{m+\bar u},
 \qquad
 p_T(\pi)\le\min\{1,\beta^{-T}q_\beta(\pi)\},
\end{equation}
and for a represented law $\widehat\mu\in\mathcal M$,
\begin{equation}\label{eq:objective-certificate}
 q_\beta(\widehat\mu)\le
 \frac{\bar u+(1-\beta)\widetilde\Jc_m(\widehat\mu)+\varepsilon_r}{m+\bar u}.
\end{equation}
No optimization guarantee is needed.
\end{proposition}

After failure, every clipped reward is $-m$, so the nonnegative cost $\bar u-r_m$ accumulates $(m+\bar u)\beta^\tau/(1-\beta)$ from failure onward. This observation is the proof (Appendix~\ref{app:performance-proofs}). To use the certificate, we need a valid value bound or, as in Sect.~\ref{sec:validation}, independent data. The geometric risk is the probability of failure before an independent geometric time, so a small $q_\beta$ does not guarantee a small probability of ever failing. Because of the factor $\beta^{-T}$, the finite-horizon certificate can be uninformative, although it remains valid. Under the interior barrier, a reserve close to zero before failure is also penalized, which never happens under a failure-only price.

\begin{remark}[The geometric risk as a risk measure of the failure time]\label{rem:failure-time}
With $\theta=\log(1/\beta)$, $q_\beta=\E e^{-\theta\tau}$ with $e^{-\theta\infty}=0$, so the geometric risk is a risk measure of the failure time that weights early failures exponentially more than late ones. It bounds the conditional value-at-risk of the failure time on the lower tail, since early failures are the bad outcomes. For every $\varepsilon\in(0,1]$, the mean failure time over the earliest $\varepsilon$-fraction of trajectories is at least $\theta^{-1}\log(\varepsilon/q_\beta)$, where a trajectory that never fails counts as failing at $+\infty$ (proved in Appendix~\ref{app:performance-proofs}). In the notation of Sect.~\ref{sec:when} this mean is $-\mathrm{AV@R}_\varepsilon(-\tau)$. A price on the geometric risk therefore bounds it from below at every fraction at once, and at $\varepsilon=p_T$ the bound gives $p_T\le\beta^{-T}q_\beta$, the second bound of \eqref{eq:risk-horizon-bound}. By contrast, a bound on $p_T$ at one horizon never gives more than $T+1$, since it does not constrain failures after $T$. Figure~E.3 of Online Resource~1 illustrates both bounds.
\end{remark}

\subsection{Utility at the attained risk level}\label{sec:performance}

Assume in this subsection that $-u_-\le u\le\bar u$, $u_-\ge0$,
and $m\ge u_-$. The \emph{stopped utility} and the \emph{interior
loss} are
\begin{align}
 U(\pi)&=\E_\pi\sum_{t<\tau}\beta^t u(s_t,a_t),
 \nonumber\\
 \mathcal I_m(\pi)&=\E_\pi\sum_{t<\tau}\beta^t
           \min\{B(z_{t+1}),u(s_t,a_t)+m\}.
 \label{eq:interior-loss}
\end{align}
Unlike the continuing utility $V(\pi)$, the stopped utility excludes
the failing action. The interior loss is nonnegative
and at most $(m+\bar u)/(1-\beta)$, even for an infinite barrier.
We split the discounted sum of clipped rewards at the failure time
and obtain the identity
\begin{equation}\label{eq:exact-risk-scalarization}
 V_m(\pi)=U(\pi)-\mathcal I_m(\pi)
                    -\frac{m}{1-\beta}q_\beta(\pi).
\end{equation}
The clipped value is thus the utility earned before failure, less the interior loss and less the discounted price $m/(1-\beta)$ times the geometric risk. A learner that raises $V_m$ trades utility against this reserve and against failure risk at this fixed rate, and the three terms are what the results of this section charge separately. Substituting the identity in the first bound of Eq.~\eqref{eq:risk-horizon-bound} shows that the right side of that bound exceeds $q_\beta(\pi)$ by exactly $[\bar u(1-q_\beta(\pi))-(1-\beta)(U(\pi)-\mathcal I_m(\pi))]/(m+\bar u)$, whose numerator is the utility per period that the policy does not earn relative to $\bar u$. The certificate is therefore close to the geometric risk when the price is large against $\bar u$, and its looseness at a moderate price is a property of the bound and not of the policy.
\begin{example}[An asset under a wear allowance]\label{ex:running}
An asset earns revenue in proportion to its operating rate $a\in[0,1]$, at a unit price that increases with a market price index $s\in[-1,1]$, and each unit of throughput
consumes a random part of a wear allowance $b=1$,
\begin{equation}\label{eq:running-example}
 u(s,a)=\frac{(1+s/2)\,a}{3/2},\qquad
 g(a,\xi)=0.1\,a\,\xi,\qquad
 s'=0.8\,s+0.2\,w ,
\end{equation}
with $w$ uniform on $[-1,1]$ and a gamma wear shock $\xi$ of mean one
and shape $1/2$. Exhausting the allowance before the next maintenance
access is a breakdown, after which each period of downtime costs $m$.
Access arrives with probability $1-\beta=0.05$ per period, so the
geometric risk is the probability of a breakdown before access. The
wear shock is unbounded, so standing still is the only policy that
never fails.
\end{example}

\begin{figure}[tb]
 \centering
 \includegraphics[width=0.9\linewidth]{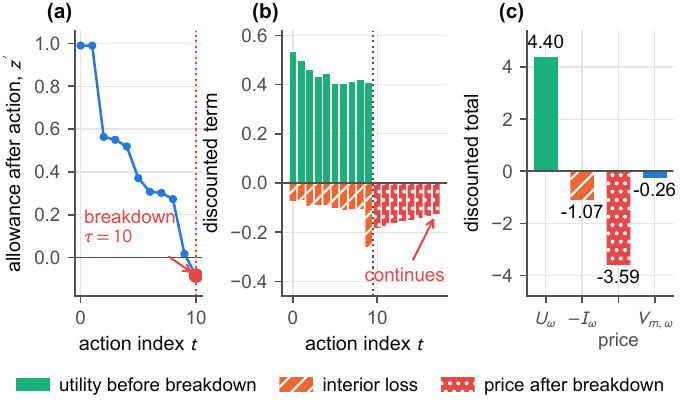}
 \caption{The return decomposition on a path of Example~\ref{ex:running}, with constant rate $a=0.8$, $b=1$,
 $\beta=0.95$, $B(z')=0.1\log(1+1/z')$, and $m=0.3$.
 (a) Random wear exhausts the allowance at action $\tau=10$, before
 maintenance access on the illustrated realization. (b) Discounted
 revenue before breakdown, interior loss and the price charged from breakdown onward. (c) Their exact pathwise totals, where
 $d=1-\beta$; the infinite sum after breakdown is evaluated in closed form.
 Subscript $\omega$ denotes this realization; $U$, $\mathcal I_m$
 and $V_m$ in Eq.~\eqref{eq:exact-risk-scalarization} are expectations.
 The path is stored in the replication repository. The figure illustrates the identity \eqref{eq:exact-risk-scalarization} and does not show a trained policy.}
 \label{fig:identity}
\end{figure}

Figure~\ref{fig:identity} shows the identity on one realized wear and price path of Example~\ref{ex:running}. In the following comparison, the losses are charged relative to a represented
comparison policy whose geometric risk is at most the comparison level. We prove the results of this section in Appendix~\ref{app:performance-proofs}.

\begin{theorem}[Matched-risk performance comparison]\label{thm:matched-risk}
Suppose \eqref{eq:risk-inexact-optimum} holds. For every
$\bar\mu\in\mathcal M$ and $\alpha\in[q_\beta(\bar\mu),1]$,
\begin{equation}\label{eq:matched-stopped}
 V(\pi_{\bar\mu})-V(\pi_{\widehat\mu})
 \le\mathcal I_m(\pi_{\bar\mu})+\mathcal R(\bar\mu)+\delta
       +\frac{2\varepsilon_r}{1-\beta}
       +\frac{m[\alpha-q_\beta(\widehat\mu)]}{1-\beta}
       +\frac{\bar u\,\alpha+u_-q_\beta(\widehat\mu)}{1-\beta}.
\end{equation}
Here $\mathcal I_m(\pi_{\bar\mu})+\mathcal R(\bar\mu)$ is the cost charged to the comparison law, namely its interior loss and its regularization cost, and $\delta+2\varepsilon_r/(1-\beta)$ is the optimization error with the reward approximation. The term with $m$ is the difference $\alpha-q_\beta(\widehat\mu)$ charged at the price, and the last term is the utility after failure. Without the last term, \eqref{eq:matched-stopped} bounds $U(\pi_{\bar\mu})-U(\pi_{\widehat\mu})$. The same conclusions hold under comparison with one specified
$\bar\mu$ instead of the infimum.
\end{theorem}

The theorem bounds the utility that can be given up by training with the clipped barrier in place of deploying a represented comparison policy. In \eqref{eq:matched-stopped}, only the comparison policy is assumed to meet the geometric level $\alpha$. The last term bounds the contribution after failure in the comparison of continuing utilities $V$. At $\alpha=q_\beta(\widehat\mu)$, the term proportional to $m$ vanishes for every comparison law of no greater risk. When both policies meet the geometric level $\alpha$, this term is the part $\alpha-q_\beta(\widehat\mu)$ of the level that the learned policy does not use, charged at the price, so excessive conservatism costs utility. Section~\ref{sec:wear} compares estimated utility and risk with a grid benchmark. Section~\ref{sec:prefer} discusses how a prescribed risk limit changes the choice of formulation. When the prior, with $\Hc(\Gamma)=0$, is the comparison law and $\alpha=q_\beta(\Gamma)$, the bound for the stopped utility still contains $m[q_\beta(\Gamma)-q_\beta(\widehat\mu)]/(1-\beta)$. This term is positive whenever the learned policy has the lower risk. The bound for preserving the prior's utility must then account for the risk reduction as well as clipped-value improvement.

\subsection{The fixed price of failure risk}\label{sec:fixed-price}

In identity \eqref{eq:exact-risk-scalarization}, each unit of
geometric risk is charged the discounted total of the price, $m/(1-\beta)$. In
the language of constrained Markov decision processes, $V_m$ is the
Lagrangian of the geometric-risk problem with the multiplier held fixed at $\nu=m/(1-\beta)$ \citep{altman1999}, less the
interior loss and up to an additive constant. With the clipped value, failure risk is thus priced rather than
held to a prescribed level. In Sect.~\ref{sec:prefer} we translate this into conditions for choosing a method. A policy that maximizes utility less a fixed multiple of the risk is optimal among the policies of no greater risk, since any such policy with higher utility would also have the higher penalized value. This sufficiency argument for penalized formulations, applied to the identity \eqref{eq:exact-risk-scalarization}, gives the following.

\begin{proposition}[Fixed price of failure risk]\label{prop:fixed-price}
Let $-u_-\le u\le\bar u$, $m\ge u_-$, $\mathcal C\subseteq\Pi$ and
$\delta\ge0$, and let $\widehat\pi\in\mathcal C$ satisfy
$V_m(\widehat\pi)\ge\sup_{\mathcal C}V_m-\delta$.
\begin{enumerate}
\item[(i)] \emph{Optimality at the attained risk.} Every
$\pi\in\mathcal C$ with $q_\beta(\pi)\le q_\beta(\widehat\pi)$ has
$U(\pi)-\mathcal I_m(\pi)\le U(\widehat\pi)-\mathcal I_m(\widehat\pi)+\delta$.
\item[(ii)] \emph{Price--risk bound.} For every $\pi_0\in\mathcal C$,
\begin{equation}\label{eq:price-risk-law}
 q_\beta(\widehat\pi)\le q_\beta(\pi_0)+\frac{1-\beta}{m}
 \Bigl[\sup_{\mathcal C}U-U(\pi_0)+\mathcal I_m(\pi_0)+\delta\Bigr].
\end{equation}
The bracket is the potential utility gain over $\pi_0$, the interior loss of $\pi_0$ and the optimization error. The bound also holds with $m+\bar u$ in place of $m$ and $\bar u\,(1-q_\beta(\pi_0))/(1-\beta)$, the most that $\pi_0$ could earn before its own failure, in place of $\sup_{\mathcal C}U$. This second form requires no knowledge of the class.
In particular, if $\pi_0$ has a finite barrier cost $\E_{\pi_0}\sum_{t<\tau}\beta^tB(z_{t+1})$, which bounds $\mathcal I_m(\pi_0)$ for every $m$, then $q_\beta(\widehat\pi)-q_\beta(\pi_0)$ is at most of order $1/m$ as the price grows at a fixed optimization error.
\item[(iii)] \emph{Monotonicity in the price.} If $u_-\le m_1<m_2$ and
$\pi_i\in\mathcal C$ maximizes $V_{m_i}$ over $\mathcal C$ up to
$\delta_i\ge0$, then
\begin{equation}\label{eq:price-monotone}
 q_\beta(\pi_2)\le q_\beta(\pi_1)
 +(1-\beta)\frac{\mathcal I_{m_2}(\pi_1)-\mathcal I_{m_1}(\pi_1)}{m_2-m_1}
 +(1-\beta)\frac{\delta_1+\delta_2}{m_2-m_1}.
\end{equation}
The second term on the right is $1-\beta$ times the slope of the interior loss of $\pi_1$ in the price, and the last term comes from the optimization errors. For every policy $\pi$, the difference quotient
$[\mathcal I_{m_2}(\pi)-\mathcal I_{m_1}(\pi)]/(m_2-m_1)$ lies between the discounted expected numbers of
periods before failure with $B(z_{t+1})>u(s_t,a_t)+m_2$ and with $B(z_{t+1})>u(s_t,a_t)+m_1$, the
periods in which the reward is clipped.
\item[(iv)] \emph{Learned laws.} If instead $\widehat\mu\in\mathcal M$
satisfies Eq.~\eqref{eq:risk-inexact-optimum}, both bounds of (ii) hold for
every comparison law $\bar\mu\in\mathcal M$, with
$\widehat\pi=\pi_{\widehat\mu}$, $\pi_0=\pi_{\bar\mu}$,
$\mathcal C=\{\pi_\mu:\mu\in\mathcal M\}$ and
$\delta'=\delta+\mathcal R(\bar\mu)+2\varepsilon_r/(1-\beta)$ in place
of $\delta$. Multiplied by $\beta^{-T}$, they bound $p_T(\widehat\mu)$.
\end{enumerate}
\end{proposition}

The proof (Appendix~\ref{app:performance-proofs}) uses only
\eqref{eq:exact-risk-scalarization} and, for (iv), the comparison of
the objectives at $\widehat\mu$ and $\bar\mu$, so parts (i) to (iii) also hold,
with $\mathcal I_m\equiv0$, for the objective $U-\nu q_\beta$ with a fixed multiplier $\nu>0$. We call $\pi_0$ in \eqref{eq:price-risk-law} the \emph{comparison policy}. With a comparison policy of zero risk and finite barrier cost, the attained risk is at most of order $1/m$. In~(iv) the bracket also contains the regularization costs, the optimization error and the reward approximation, so raising $m$ need not help when these grow with the price or when the risks of the represented class have a positive lower bound (Remark~\ref{rem:bias}). By~(iii), the attained risk can rise with a higher price only by the optimization errors and by the clipped periods before failure, which a failure-only price does not have. A higher price need not change the returned policy at all. Since $r_{m_2}\le r_{m_1}$, the policy $\pi_1$ of (iii) satisfies
\begin{equation}\label{eq:price-gap-at-stake}
 \sup_{\mathcal C}V_{m_2}-V_{m_2}(\pi_1)\le\delta_1
 +\frac{m_2-m_1}{1-\beta}q_\beta(\pi_1)
 +\mathcal I_{m_2}(\pi_1)-\mathcal I_{m_1}(\pi_1),
\end{equation}
as we show in Appendix~\ref{app:performance-proofs}. It is thus already optimal at $m_2$ up to the price gap times its risk and the increase of its interior loss, and a learner whose optimization error exceeds this amount can return $\pi_1$ at either price. The effect of a price is therefore in proportion to the risk that exists without it (Sect.~\ref{sec:lunar}).

The price is also a shadow price of risk. For the policy $\widehat\pi$ of Proposition~\ref{prop:fixed-price}, put $\widehat q=q_\beta(\widehat\pi)$ and $\nu=m/(1-\beta)$, and let $\mathcal F_m(\alpha)=\sup\{U(\pi)-\mathcal I_m(\pi):\pi\in\mathcal C,\ q_\beta(\pi)\le\alpha\}$ be the frontier of utility net of interior loss against geometric risk over $\mathcal C$, equal to $-\infty$ when no policy of the class has risk at most $\alpha$. Then $\mathcal F_m(\widehat q)\ge U(\widehat\pi)-\mathcal I_m(\widehat\pi)$ and, for every $\alpha\in[0,1]$,
\begin{equation}\label{eq:price-supports-frontier}
 \mathcal F_m(\alpha)\le U(\widehat\pi)-\mathcal I_m(\widehat\pi)+\delta+\nu\,(\alpha-\widehat q),
\end{equation}
so for $\delta=0$ the discounted price $\nu$ is a supergradient of $\mathcal F_m$ at $\widehat q$ (Appendix~\ref{app:performance-proofs}). By~(i), every fixed-price optimum lies on this frontier (Fig.~\ref{fig:logic}(b)). Over a represented class, the frontier need not be concave, so only the frontier points supported by some slope can be reached with a price.

\begin{figure}[tb]
 \centering
 \includegraphics[width=0.9\linewidth]{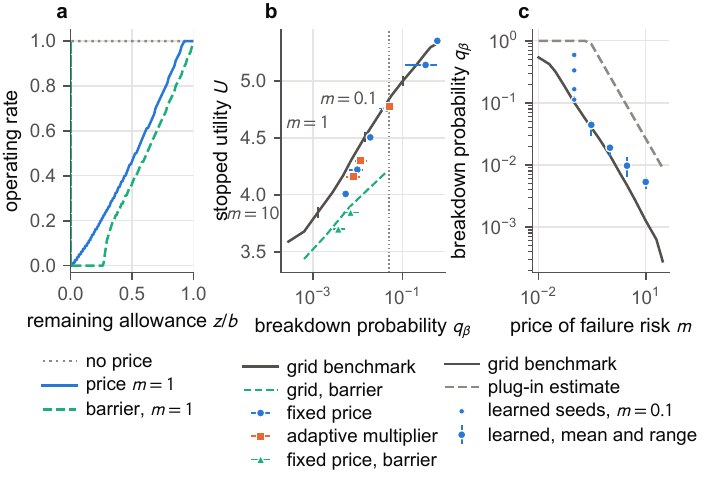}
 \caption{Price and operating policy in the wear-allowance model ($\beta=0.95$, $b=1$, and $B(z')=0.1\log(1+b/z')$ for the interior barrier).
 (a) Grid-based policies as the allowance shrinks.
 (b) Interpolated utility--risk curves for grid benchmarks and estimated outcomes for fixed-price and adaptive-multiplier PPO policies, with tolerance $0.05$.
 (c) Estimated geometric risk against price. The dashed curve evaluates the price--risk expression for an exact optimizer and a standing-still comparison policy, substituting the simulated no-price benchmark value for $\sup_{\mathcal C}U$.}
 \label{fig:wear-frontier}
\end{figure}

Figure~\ref{fig:wear-frontier} compares the learned policies with grid-based benchmarks in Example~\ref{ex:running}. At positive prices the learned policies have lower estimated breakdown risk and stopped utility than the policy trained without a price. The adaptive-multiplier and failure-only runs are close to the same benchmark curve. Online Resource~1 gives two further illustrations. Its Appendix~F.5 draws Theorem~\ref{thm:matched-risk} and Proposition~\ref{prop:fixed-price} on Example~\ref{ex:running} for a class of simple rules, and its Example~E.1 shows a frontier with a point that no price supports.

A prescribed tolerance can be unattainable. If
$\underline q=\inf_{\pi\in\mathcal C}q_\beta(\pi)>\alpha$, then $\sup_{\pi\in\mathcal C}[U(\pi)-\nu(q_\beta(\pi)-\alpha)]\le\sup_{\mathcal C}U-\nu(\underline q-\alpha)$, which tends to $-\infty$ as $\nu\to\infty$,
so there is no feasible policy and the exact dual has no finite minimizer. The clipped value is bounded at each finite price.

Suppose that the prices grow without bound while the interior loss of a fixed comparison policy and the optimization error grow more slowly than the price. By~(ii), the attained geometric risk then eventually falls below every level above the risk of the comparison policy, so a tolerance strictly above the risk of some represented policy is eventually met. The condition on the optimization error is an assumption about the learner, not a consequence of the theory.

\subsection{Action regularization and reference risk}
\label{subsec:action-risk}

The action relative entropy accumulates into a relative entropy between
trajectory laws, and in this way a local policy penalty is linked to the failure risks
$p_T$ and $q_\beta$. Let $p_T(\rho)$
and $q_\beta(\rho)$ be the risks when actions are drawn from $\rho$, and
write $\mathrm{kl}(p\|q)=p\log(p/q)+(1-p)\log((1-p)/(1-q))$ for
$p,q\in[0,1]$, with the usual conventions at the boundary.

\begin{proposition}[Action regularization and failure risk]
\label{prop:action-risk-bridge}
For every represented law $\mu$ and integer $T\ge0$,
\begin{equation}\label{eq:action-risk-geometric}
 \mathrm{kl}(p_T(\mu)\|p_T(\rho))\le\beta^{-T}\mathcal A(\mu),
 \qquad
 \mathrm{kl}(q_\beta(\mu)\|q_\beta(\rho))\le\mathcal A(\mu).
\end{equation}
Since $\mathrm{kl}(p\|q)\ge2(p-q)^2$, the finite-horizon risk of
$\pi_\mu$ lies within $\sqrt{\mathcal A(\mu)/(2\beta^T)}$ of that of the action
reference, and its geometric risk within $\sqrt{\mathcal A(\mu)/2}$.
The bounds also hold for other policy kernels in the same model with finite statewise relative entropy to $\rho$. They use the achieved action relative entropy and require no parameter regularization. We give the proof in Appendix~\ref{app:performance-proofs}.
\end{proposition}

With a small achieved $\mathcal A$, the failure risk stays close to that of the reference. The risk of the reference therefore helps determine whether an action penalty is appropriate. A high-risk reference is a reason against strong
anchoring. The bounds can be loose as numerical certificates while
still explaining the behavioral bias. A small achieved $\mathcal A$ does not follow from
increasing $\eta$ alone.

\section{Convergence under general convex regularization}\label{sec:flow}

We give sufficient conditions for convergence of the population gradient flow and explain what each regularizer contributes. The certificates of Sect.~\ref{sec:risk} hold for whatever policy an optimizer returns, but convergence of training is not automatic, because the clipped value is not convex in the law of the network's units. The condition we derive is a \emph{curvature budget}. A parameter regularizer that is convex along motions of the units supplies curvature, and once this supply exceeds the curvature that the objective lacks, the gradient flow converges exponentially. A larger strength can increase the weight of the prior or the reference in the objective. We assess the guaranteed rate together with the resulting regularization cost (Sect.~\ref{subsec:prior}).

Theorem~\ref{thm:flow} extends the relative-entropy analysis of mean-field policy gradients of \citet[Theorem~2.12]{leahy2022} to the clipped value on the budget-augmented state and to every parameter regularizer convex along such motions, including relative entropy with a displacement-convex integrand and convex potentials such as weight decay. Under relative entropy, \citet{lascu2025} obtain convergence above a strength threshold by means of a best-response flow, and the only analysis with a generalized entropy that we know of linearizes the potential \citep{akiyama2024}. Throughout, the represented class, the action reference, the price $m$ and the strengths $\eta$ and $\kappa$ stay fixed.

The objects are those of Sect.~\ref{sec:population-objective}, and the learning objective is $\Jc_m$ of Eq.~\eqref{eq:training-objective}, extended by $+\infty$ outside $\mathcal M$. For $\eta=0$ the action term is absent
and $C_{\mathcal A}=0$.

\begin{assumption}[Features and prior]\label{ass:flow}
\begin{enumerate}
\item[(a)] The feature $\psi(y,a,x)$ of Eq.~\eqref{eq:softmax} is jointly measurable and twice continuously
differentiable in $x\in\R^d$, with $|\psi|\le L_0$,
$|\nabla_x\psi|\le L_1$ and $\|\nabla_x^2\psi\|_{\mathrm{op}}\le L_2$
uniformly in $(y,a,x)$, for finite constants $L_0,L_1,L_2$.
The Hessian is uniformly continuous in $x$, uniformly over $(y,a)$:
$\sup_{y,a,|x-x'|\le h}\|\nabla_x^2\psi(y,a,x)-\nabla_x^2\psi(y,a,x')\|_{\rm op}\to0$
as $h\downarrow0$. The uniform Taylor remainders used in
Appendix~\ref{app:curvature} follow from this; they do not follow from pointwise $C^2$
regularity and a bounded Hessian alone.
\item[(b)] The prior has a density $\Gamma(dx)=e^{-W(x)}dx$
with $W\in C^2(\R^d)$ and $\nabla^2W\succeq\lambda_W I$ for some
$\lambda_W>0$; the additive constant in $W$ normalizes the density. A
Gaussian $N(x_0,\sigma^2I)$ has $\lambda_W=\sigma^{-2}$. This
strong log-concavity is what makes the relative entropy to the prior
convex under transport, with curvature $\lambda_W$
(Proposition~\ref{prop:generators}).
\end{enumerate}
\end{assumption}

Part~(a) is all that the convergence theorem requires of the policy class. The transition kernel may be merely Borel, and the clipped
reward need not be differentiable in the state or the action. The proof uses only bounds on how fast the policy and its relative entropy to the action reference change when the units move
(Appendix~\ref{app:curvature}), which the softmax class satisfies with
constants computed from $L_0,L_1,L_2$ and which can be verified for another policy class.

The relevant notion of convexity is convexity under transport. The curvature counted in the budget
is convexity of $\Hc$ when every unit moves on a straight line from one
configuration to another, not convexity under mixing of laws. For the
relative entropy to a log-concave prior this is displacement
convexity \citep[Chapter~9]{ags2005}. Formally, convexity on $\Ptwo$, with
the Wasserstein-2 distance $\Wtwo$, is understood along
\emph{generalized geodesics} \citep[Definition~9.2.4]{ags2005}. Given a
base law $\Xi\in\Ptwo$ and two laws $\mu_0,\mu_1$, take a plan
$\gamma$ on $\R^d\times\R^d\times\R^d$ whose marginals on the first two
and on the first and third coordinates are optimal couplings of
$(\Xi,\mu_0)$ and $(\Xi,\mu_1)$. Let $\mu_\vartheta$ be the law
of $(1-\vartheta)x_2+\vartheta x_3$ under $\gamma$, $\vartheta\in[0,1]$. A
functional $\phi$ on $\Ptwo$ is \emph{$\lambda$-convex along
generalized geodesics} if for every such $\Xi,\mu_0,\mu_1$ at which
$\phi$ is finite some such plan satisfies
\[
 \phi(\mu_\vartheta)\le(1-\vartheta)\phi(\mu_0)+\vartheta\phi(\mu_1)
 -\tfrac{\lambda}{2}\vartheta(1-\vartheta)W_\gamma^2(\mu_0,\mu_1),
\]
where $W_\gamma^2(\mu_0,\mu_1)=\int|x_2-x_3|^2d\gamma\ge\Wtwo^2(\mu_0,\mu_1)$.

\begin{definition}[Admissible parameter regularizer]
A functional $\Hc:\Ptwo\to[0,\infty]$ is \emph{admissible with
curvature} $\lambda_{\Hc}\in\R$ if it is proper, narrowly lower
semicontinuous and $\lambda_{\Hc}$-convex along generalized
geodesics; narrow lower semicontinuity implies lower semicontinuity
in $\Wtwo$, which is what the theory of gradient flows uses.
\end{definition}

\begin{proposition}[Admissible families]\label{prop:generators}
Let $\Gamma(dx)=e^{-W(x)}dx$ be the prior of
Assumption~\ref{ass:flow}(b). Let $G:[0,\infty)\to[0,\infty)$ be
convex with $G(0)=0$ and $G(v)/v\to\infty$ as $v\to\infty$, and
suppose that
\begin{equation}\label{eq:mccann}
 s\longmapsto e^{s}G(e^{-s})
 \quad\text{is convex and nonincreasing on }\R .
\end{equation}
For $\omega\ge0$ define, for $\mu=\varrho\,\Gamma$ with density $\varrho$ and
$+\infty$ otherwise,
\begin{equation}\label{eq:general-divergence}
 \Hc_{\omega,G}(\mu)=\KL(\mu\|\Gamma)
 +\omega\Bigl[\int G(\varrho)\,d\Gamma-G(1)\Bigr],
\end{equation}
and call $G$ the \emph{integrand} of $\Hc_{\omega,G}$. Then
$\Hc_{\omega,G}$ is admissible with curvature $\lambda_W$,
$\Hc_{\omega,G}(\Gamma)=0$, and
$\Hc_{\omega,G}\ge\KL(\cdot\|\Gamma)$; the case $\omega=0$ is
relative entropy alone. All hypotheses hold for the power integrands
$G(v)=v^r/(r-1)$, $r>1$, and are preserved by finite nonnegative
combinations that are not identically zero. The choice $r=2$ gives
\begin{equation}\label{eq:flow-divergence}
 \Hc_\omega(\mu)=\KL(\mu\|\Gamma)+\omega\chi^2(\mu\|\Gamma)
 =\int\bigl[\varrho\log\varrho+\omega(\varrho-1)^2\bigr]d\Gamma,
 \qquad \mu=\varrho\,\Gamma.
\end{equation}
For a continuous $\lambda_P$-convex $P:\R^d\to[0,\infty)$, the potential $\Hc_P(\mu)=\int P\,d\mu$ is admissible with curvature $\lambda_P$ and finite on every finitely supported law.
\end{proposition}

For $\omega>0$, finiteness in Eq.~\eqref{eq:flow-divergence} requires
$\varrho\in L^2(\Gamma)$. For $\omega=0$ the regularizer is
relative entropy alone on its full finite-entropy domain. Condition \eqref{eq:mccann} is the dimension-free
displacement-convexity condition of \citet[Remark~9.3.10]{ags2005}, stated there for $s>0$.
The curvature does not depend on the integrand, so the rate and the threshold
are the same for every integrand, whereas the bias $\kappa\Hc(\bar\mu)$
is not. Weight decay is a potential
with curvature one, and it is finite on the empirical law
of a finite network. Potentials are included here as the finite-network counterpart of the quadratic part of a Gaussian prior
(Sect.~\ref{subsec:prior}). A divergence of Proposition~\ref{prop:generators} with strength $\kappa_1$ plus a potential with strength $\kappa_2$ is admissible with curvature $\kappa_1\lambda_W+\kappa_2\lambda_P$, so each of the two can be chosen for what it controls. Figure~E.4 of Online Resource~1 illustrates the condition and the convexity along generalized geodesics.

We first bound the curvature of the value and action terms. Upper bounds on the curvature deficits of
the clipped value and of the action relative entropy are, for the
softmax class,
\begin{equation}\label{eq:flow-curvature-constant}
 C_m=(m+\bar u)c_\psi,\qquad
 c_\psi=\frac{L_2}{(1-\beta)^2}+\frac{L_1^2}{(1-\beta)^3},
\end{equation}
with $C_{\mathcal A}$ given by Eq.~\eqref{eq:action-curvature-constants}
in Appendix~\ref{app:curvature}. These bounds are sufficient, not necessarily
sharp, and they account for the change of the state visitation with the
policy. The curvature budget is
\[
 \Lambda_m=\kappa\lambda_{\Hc}-C_m-\eta C_{\mathcal A},
\]
the supply less the deficit. The learning dynamics is the gradient
flow of $\Jc_m$ in the Wasserstein space, in the sense of the \emph{evolution variational inequality}, that is, a locally absolutely continuous curve $(\mu_\ell)_{\ell>0}$ in $(\Ptwo,\Wtwo)$, continuous
at its initial law, such that for every $\mu'$ of finite objective and
almost every $\ell>0$,
\begin{equation}\label{eq:flow-evi-main}
 \frac12\frac{d}{d\ell}\Wtwo^2(\mu_\ell,\mu')
 +\frac{\Lambda_m}{2}\Wtwo^2(\mu_\ell,\mu')
 \le\Jc_m(\mu')-\Jc_m(\mu_\ell).
\end{equation}

\begin{theorem}[Convergence of the population gradient flow]\label{thm:flow}
Under Assumption~\ref{ass:flow}(a), let $\Hc$ be admissible with
curvature $\lambda_{\Hc}$. For every $\mu_0\in\mathcal M$, \eqref{eq:training-objective} has a unique
gradient flow $(\mu_\ell)_{\ell\ge0}$ in the sense of Eq.~\eqref{eq:flow-evi-main}. Its objective is nonincreasing, and two
gradient flows from $\mu_0,\mu'_0\in\mathcal M$ satisfy
$\Wtwo(\mu_\ell,\mu'_\ell)\le e^{-\Lambda_m\ell}\Wtwo(\mu_0,\mu'_0)$
for every sign of $\Lambda_m$.
If $\Lambda_m>0$, the objective has a unique global minimizer $\mu_m^*$, and
\begin{align}
 \Wtwo(\mu_\ell,\mu_m^*)&\le
 e^{-\Lambda_m\ell}\Wtwo(\mu_0,\mu_m^*),
 \label{eq:flow-distance-rate}\\
 0\le\Jc_m(\mu_\ell)-\Jc_m(\mu_m^*)&\le E_0e^{-2\Lambda_m\ell},
 \qquad E_0=\Jc_m(\mu_0)-\Jc_m(\mu_m^*).
 \label{eq:flow-energy-rate}
\end{align}
No low-risk prior or alignment $\rho=\pi_\Gamma$ is required.
The optimum is that of the represented regularized objective, not
the chance-constrained or almost-sure constrained optimum.
\end{theorem}

In Appendix~\ref{app:optimization-proofs} we verify the semiconvexity of
the clipped value and the action relative entropy along every coupling
interpolation and apply the theory of \citet[Theorem~11.2.1]{ags2005}.

\begin{figure}[tb]
 \centering
 \includegraphics[width=0.8\linewidth]{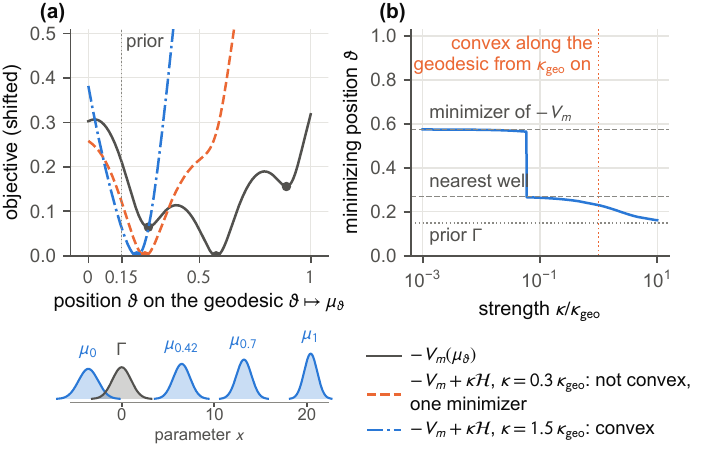}
 \caption{The curvature budget of Theorem~\ref{thm:flow} along a geodesic of Gaussian parameter laws. A unit has a real parameter $x$, the prior is $\Gamma=N(0,1)$, the parameter regularizer is $\kappa\Hc=\kappa\KL(\cdot\|\Gamma)$ with $\lambda_{\Hc}=1$, and $\eta=0$. The policy of a law $\mu$ depends on $\mu$ through the number $\int\varphi\,d\mu$ alone, and the clipped value increases in this number, as in the analytical example of Sect.~\ref{sec:flow-test}. The feature $\varphi$ is smooth and has three peaks of different height on the range shown (Appendix~F.6 of Online Resource~1). The law $\mu_\vartheta$ at the position $\vartheta\in[0,1]$ is Gaussian with mean and standard deviation linear in $\vartheta$. These laws form a generalized geodesic, which passes through the prior at $\vartheta=0.15$, and the strip below panel~(a) shows the densities of $\mu_0$, $\Gamma$, $\mu_{0.42}$, $\mu_{0.7}$ and $\mu_1$ on the scale of the parameter $x$. We write $\kappa_{\mathrm{geo}}$ for the smallest strength at which $-V_m+\kappa\Hc$ is convex along this geodesic. It is at most $C_m$, the strength above which $\Lambda_m>0$. Curves are shifted to a minimum of zero.
 (a)~The negative clipped value $-V_m(\mu_\vartheta)$ has three local minimizers of different depth, which we call wells. At $1.5\,\kappa_{\mathrm{geo}}$ the sum is convex, and its minimizer lies between the prior and the nearest well. At $0.3\,\kappa_{\mathrm{geo}}$ the sum is not convex and has one minimizer.
 (b)~The minimizer among the laws of the geodesic against the strength. As the strength grows, it jumps from the deepest well to the nearest well, stays near that well up to $\kappa_{\mathrm{geo}}$ and then moves to the prior. This is the trade-off between convergence and bias.}
 \label{fig:curvature}
\end{figure}

We illustrate the mechanism in Fig.~\ref{fig:curvature} along one geodesic of Gaussian laws. Above the threshold the regularized objective has one minimizer and the gradient flow contracts toward it. At weak strengths the objective of the figure has several local minimizers along the geodesic, and the minimizer in panel~(b) jumps from the deepest one to the one nearest to the prior as the strength grows. Once the objective is convex along the geodesic, its minimizer lies between the prior and this nearest local minimizer, so the quality of the solution then depends on the prior. The cost is a bias toward the prior, since the rate $\Lambda_m$ and the bias terms $\kappa\Hc(\bar\mu)$ of Proposition~\ref{prop:fixed-price}(iv) and Theorem~\ref{thm:matched-risk} are both linear in $\kappa$. For $\lambda_{\Hc}>0$, the strength $(C_m+\eta C_{\mathcal A})/\lambda_{\Hc}$ above which $\Lambda_m>0$ is the \emph{threshold} of the theorem. It holds uniformly over all parameter laws, and in the simulations of Sect.~\ref{sec:flow-test} fitted gap curves also decay at strengths far below it. On the action simplex the regularized greedy policy is unique under a strongly convex action regularizer \citep[Proposition~1]{geist2019}, but convexity in the action law does not transfer through the nonlinear map $\mu\mapsto\pi_\mu$. In Theorem~\ref{thm:flow} the action relative entropy therefore contributes no curvature and is charged $\eta C_{\mathcal A}$. It acts as a reward term that holds the policy near the action reference (Proposition~\ref{prop:action-risk-bridge}), and an entropy bonus corresponds to a uniform action reference when the policy uses that reference measure.

\begin{corollary}[Optimization error and risk along the gradient flow]\label{thm:combined}
Assume $-u_-\le u\le\bar u$, $m\ge u_-$, and the setting of Theorem~\ref{thm:flow}
with $\Lambda_m>0$. For every comparison law $\bar\mu\in\mathcal M$
and learning time $\ell\ge0$, both bounds of
Proposition~\ref{prop:fixed-price}(iv) hold for
$\widehat\mu=\mu_\ell$ with $\varepsilon_r=0$ and
$\delta'=\mathcal R(\bar\mu)+E_0e^{-2\Lambda_m\ell}$; in the form
\eqref{eq:price-risk-law}, with $\mathcal C$ the represented class,
\begin{equation}\label{eq:combinedrisk}
 q_\beta(\mu_\ell)\le q_\beta(\bar\mu)
 +\frac{1-\beta}{m}
 \Bigl[\sup_{\mathcal C}U-U(\pi_{\bar\mu})+\mathcal I_m(\pi_{\bar\mu})
       +\mathcal R(\bar\mu)+E_0e^{-2\Lambda_m\ell}\Bigr].
\end{equation}
Dividing by $\beta^T$ and capping at one bounds $p_T(\mu_\ell)$. The
comparison law may have positive risk and need not be the prior.
\end{corollary}

We prove the corollary in Appendix~\ref{app:optimization-proofs}. The
curvature the regularizer must supply grows linearly with the price,
so $\kappa$ cannot vanish as $m$ grows if $c_\psi>0$. This limitation lies in the sufficient estimates and is not a lower bound on risk.

Theorem~\ref{thm:flow} holds the strength fixed. The next result concerns a strength that decreases to zero, at a fixed price $m$ and a fixed weight $\eta\ge0$. We write $\Jc_m^{0}=-V_m+\eta\mathcal A$ and $\Jc_m^{\kappa}=\Jc_m^{0}+\kappa\Hc$ for the objective \eqref{eq:training-objective} at the strength $\kappa$. Functionals $f_j$ on $(\Ptwo,\Wtwo)$ \emph{epi-converge} to $f$ if $\liminf_jf_j(\mu_j)\ge f(\mu)$ whenever $\mu_j\to\mu$, and every $\mu$ is the limit of a sequence with $\limsup_jf_j(\mu_j)\le f(\mu)$.

\begin{proposition}[Decreasing strength]\label{prop:decreasing-strength}
Let Assumption~\ref{ass:flow}(a) hold, and let $\Hc$ be admissible with $\mathcal M$ dense in $(\Ptwo,\Wtwo)$, as it is for the regularizers of Proposition~\ref{prop:generators}. Along every sequence $\kappa_j\downarrow0$, the objectives $\Jc_m^{\kappa_j}$ epi-converge to $\Jc_m^{0}$. If $\Jc_m^{\kappa_j}(\mu_j)\le\inf\Jc_m^{\kappa_j}+\delta_j$ with $\delta_j\downarrow0$, then $\Jc_m^{0}(\mu_j)\to\inf_{\Ptwo}\Jc_m^{0}$ and $\kappa_j\Hc(\mu_j)\to0$, and every cluster point of $(\mu_j)_j$ minimizes $\Jc_m^{0}$ over $\Ptwo$.
\end{proposition}

We prove the proposition in Appendix~\ref{app:optimization-proofs}. It concerns minimizers and not the learning dynamics, and it requires no curvature. For $\eta=0$ the clipped values of the regularized minimizers approach the supremum of the clipped value. The supremum need not be attained, and then there is no cluster point, as in the analytical example of Sect.~\ref{sec:flow-test}. Two further properties follow in one line each. If $\mu^{\kappa}$ and $\mu^{\kappa'}$ minimize the objective at the strengths $\kappa<\kappa'$, adding the two minimality inequalities gives $\Hc(\mu^{\kappa})\ge\Hc(\mu^{\kappa'})$ and $\Jc_m^{0}(\mu^{\kappa})\le\Jc_m^{0}(\mu^{\kappa'})$, so for $\eta=0$ the smaller strength gives a minimizer with clipped value at least as high. If the strength is lowered in steps during training and the law follows the gradient flow of Theorem~\ref{thm:flow} at the current strength, the current objective never increases, because $\Hc\ge0$. The clipped value then never falls below its initial value by more than the initial regularization cost, and the rates of Theorem~\ref{thm:flow} hold on the steps above the threshold. Whether this gradient flow stays close to the path of minimizers below the threshold is not proved here.

\begin{remark}[Regularization cost in the risk comparison]\label{rem:bias}
The comparisons charge the regularization cost $\mathcal R(\bar\mu)=\eta\mathcal A(\bar\mu)+\kappa\Hc(\bar\mu)$ of the comparison law. In Theorem~\ref{thm:matched-risk} it enters in utility units. The risk bounds in Proposition~\ref{prop:fixed-price}(iv) multiply it by $(1-\beta)/(m+\bar u)$ or $(1-\beta)/m$. A tolerance is therefore certified when some comparison law leaves enough slack for all charged terms. If all comparison laws meeting the requirement have a positive lower bound on $\Hc$, and the strength grows proportionally to the price, the normalized parameter term can stay positive as the price grows. A failed certificate can also result from a loose bound, limited validation data or a candidate that exceeds the tolerance. The theorem gives a sufficient strength. Runs at smaller strengths can be assessed on their own optimization and validation evidence (Sect.~\ref{sec:flow-test}).
\end{remark}

\subsection{Improvement relative to the prior policy}
\label{subsec:baseline-performance}

A divergence regularizer of Proposition~\ref{prop:generators} also has a direct risk interpretation, which for $\eta>0$ requires the action reference to be the prior policy. The action reference is \emph{aligned} with the \emph{prior policy} $\pi_\Gamma$, the policy \eqref{eq:softmax} of the prior, when
\begin{equation}\label{eq:centered-baseline}
 f_\Gamma(y,a)=0,\qquad\rho(\cdot\mid y)=\pi_\Gamma(\cdot\mid y),
\end{equation}
which is achieved, with $\rho$ unchanged, by replacing the feature by its version centered at the prior, $\psi(y,a,x)-\int\psi(y,a,x')\,\Gamma(dx')$. The bound $L_0$ can then double, while the derivative bounds do not change. Alignment is an objective-design choice, since the action relative entropy changes with it. Let $\Hc$ be admissible with $\Hc(\Gamma)=0$ and $\|\mu-\Gamma\|_{\rm TV}^2\le c_{\Hc}\Hc(\mu)$, which holds with $c_{\Hc}=2$ for the divergence families of Proposition~\ref{prop:generators} by Pinsker's inequality, let $R_\Gamma=\bar u/(1-\beta)-V_m(\Gamma)$, and let $(\mu_\ell)$ be the gradient flow of Theorem~\ref{thm:flow} started at $\Gamma$ under Assumption~\ref{ass:flow}(a), with either $\eta=0$ or the alignment \eqref{eq:centered-baseline}. No sign condition on $\Lambda_m$ is needed.

\begin{proposition}[Improvement over the prior policy]
\label{prop:online-baseline}
For every $\ell\ge0$ and integer $T\ge0$,
\begin{equation}\label{eq:online-baseline-slack}
 V_m(\mu_\ell)-V_m(\Gamma)\ge\eta\mathcal A(\mu_\ell)+\kappa\Hc(\mu_\ell)\ge0,
 \qquad
 p_T(\mu_\ell)\le p_T(\Gamma)+\frac{L_0}{2}
     \sqrt{\frac{(T+1)c_{\Hc}R_\Gamma}{\kappa}},
\end{equation}
and, if $\eta>0$,
$|p_T(\mu_\ell)-p_T(\Gamma)|\le\sqrt{R_\Gamma/(2\eta\beta^T)}$; the
smaller slack applies. In particular, a tolerance $\alpha>p_T(\Gamma)$
is met by every $\mu_\ell$ once
$\kappa\ge L_0^2(T+1)c_{\Hc}R_\Gamma/[4(\alpha-p_T(\Gamma))^2]$.
\end{proposition}

Every policy on the gradient flow thus has a clipped value at least that of the prior policy and a finite-horizon risk at most that of the prior policy plus a term of order $\kappa^{-1/2}$. A lower bound on $V_m(\Gamma)$ bounds $R_\Gamma$ without reference to an optimal policy. The proposition asserts neither violation-free learning, nor strict improvement of the original utility, nor convergence of a finite-network optimizer. We prove it in Appendix~\ref{app:prior-proof}.

\subsection{The prior and the particle system of the gradient flow}
\label{subsec:prior}

The particle system of the population gradient flow shows what each regularizer does to a single unit. Let $\Hc=\kappa_1\Hc_{\omega,G}+\kappa_2\Hc_P$, a divergence and a potential of Proposition~\ref{prop:generators} with their strengths. Write the network as units with parameters $x_1,\dots,x_n$ and empirical law $\mu_n$. Let $B_1,\dots,B_n$ be independent standard Brownian motions. We write the density-dependent force formally using $\varrho=d\mu/d\Gamma$ for a law $\mu$ that has a density relative to the prior. The first variation $\frac{\delta\phi}{\delta\mu}(\mu,\cdot)$ of a functional $\phi$ at $\mu$ is the function whose integral against $\mu'-\mu$ is the derivative of $\phi(\mu+\epsilon(\mu'-\mu))$ at $\epsilon=0$. With the strengths inside $\Hc$ ($\kappa=1$), each unit moves along the gradient flow according to
\begin{equation}\label{eq:particle}
 \begin{aligned}
 dx_i={}&\Bigl[\nabla_x\frac{\delta V_m}{\delta\mu}(\mu,x_i)
 -\eta\,\nabla_x\frac{\delta\mathcal A}{\delta\mu}(\mu,x_i)\Bigr]d\ell\\
 &-\Bigl[\kappa_1\nabla W(x_i)
 +\kappa_1\omega\,\nabla\bigl[G'(\varrho)\bigr](x_i)
 +\kappa_2\nabla P(x_i)\Bigr]d\ell+\sqrt{2\kappa_1}\,dB_i .
 \end{aligned}
\end{equation}
For $\kappa_1>0$ and $\omega=0$, Eq.~\eqref{eq:particle} with $\mu=\mu_n$ is the mean-field Langevin particle system. For $\kappa_1\omega>0$ the displayed density term is a formal force for the population law, since the empirical law has no density relative to $\Gamma$. The first two terms are the policy gradients of the clipped value and of the action relative entropy with respect to the unit's parameters (Proposition~\ref{prop:first-variations}). The remaining terms are the regularizers, each with its own role. Relative entropy gives Langevin noise of variance $2\kappa_1$ per unit time on every unit. Standard learners do not inject such noise, and it allows units to leave a region where the value gradient vanishes. The prior's potential $W$ gives a deterministic drift toward the prior, which for a Gaussian prior is weight decay toward its center with strength $\kappa_1/\sigma^2$. Under the integrand term, units move out of regions where the law is crowded relative to the prior, which limits how far the law of the units departs from the prior. The potential $P$ gives a second deterministic drift, toward a chosen reference, without noise. Under relative entropy alone, the regularizer of \citet{leahy2022}, the noise and the drift are tied together by the single strength $\kappa_1$. In the general class they are separate, and the strengths add $\kappa_1\lambda_W+\kappa_2\lambda_P$ to the curvature budget while the integrand adds none (Sect.~\ref{sec:flow-test}).

Anchoring every unit quadratically to one common reference is the potential term on a finite network. The anchor of Appendix~D.5 of Online Resource~1, in which each parameter of a network is tied to its own value in an incumbent, is a different penalty. It is not a function of the law of the units, and Proposition~\ref{prop:finite-training} does not apply to it. For the entropy-only Langevin system, Sect.~\ref{sec:flow-test} compares empirical laws at several widths with a numerical approximation to the population flow. Convergence as width grows remains open here. For a common potential alone, the next proposition gives the exact correspondence with $\mu=\mu_n$.

\begin{proposition}[Finite-network training under a potential]
\label{prop:finite-training}
Let Assumption~\ref{ass:flow}(a) hold, $\eta\ge0$, and
$\Hc=\int P\,d\mu$ with $P\ge0$ $\lambda_P$-convex for some $\lambda_P\ge0$ and continuously
differentiable with a Lipschitz gradient, so that $\mathcal M=\Ptwo$ and
$\Lambda_m=\kappa\lambda_P-C_m-\eta C_{\mathcal A}$. For $n$ units
with any initial parameters, the system \eqref{eq:particle} without
noise, $\kappa_1=0$ and $\kappa_2=\kappa$, has a unique solution, and
the empirical law $\mu_n(\ell)$ of its units is the gradient flow of
Theorem~\ref{thm:flow} from $\mu_n(0)$. Hence, for the gradient flow
$(\mu_\ell)$ from any $\mu_0\in\Ptwo$ and every $\ell\ge0$,
\begin{equation}\label{eq:finite-training}
 \Wtwo(\mu_n(\ell),\mu_\ell)\le e^{-\Lambda_m\ell}\Wtwo(\mu_n(0),\mu_0),
\end{equation}
the clipped values and geometric risks of the two laws differ
by at most $(m+\bar u)L_1/(2(1-\beta)^2)$ and $L_1/(2(1-\beta))$ times
that distance, and if $\Lambda_m>0$ then
$\Wtwo(\mu_n(\ell),\mu_m^*)\le e^{-\Lambda_m\ell}\Wtwo(\mu_n(0),\mu_m^*)$,
where $\mu_m^*=\delta_{x^*}$ is a point mass and $x^*$ minimizes
$x\mapsto-V_m(\delta_x)+\eta\mathcal A(\delta_x)+\kappa P(x)$, so that the units of a network of any width converge to $x^*$.
\end{proposition}

Continuous-time gradient descent with weight
decay or an anchor to one common reference, on a network of any width and with the step scaled by the width, is thus the population gradient flow from the network's own
empirical law and inherits its contraction and rate. The units' velocity is minus an element of the objective's subdifferential at an atomic law, so the particle
curve is a gradient flow \citep{chizat2018}, and by uniqueness it coincides with the population gradient flow
(Appendix~\ref{app:finite-training-proof}).

For a Gaussian prior $\Gamma=N(x_0,\sigma^2I)$ and a law $\mu$ with a density, $\log(d\mu/d\Gamma)=\log(d\mu/dx)+|x-x_0|^2/(2\sigma^2)$ up to a constant, so relative entropy of strength $\kappa$ is a negative-entropy term of strength $\kappa$ plus quadratic weight decay toward $x_0$ of strength $\kappa/\sigma^2$. Only the quadratic part is ordinary weight decay. It stays finite on the empirical law of a finite network, whereas the entropy does not. Since $\lambda_W=\sigma^{-2}$, the rate of Theorem~\ref{thm:flow} depends on $(\kappa,\sigma)$ only through $\kappa/\sigma^2$, while the objective and its minimizer depend on both.

In the analytical example of Sect.~\ref{sec:flow-test}, the computed minimizers at moderate strengths improve on the prior policy. At the largest tested strengths the clipped-value gain is small.
In Appendix~D.5 of Online Resource~1 we test an analogue on a learned controller,
namely quadratic anchoring of PPO's parameters to those of an incumbent.

\section{Finite-network deployment and validation}
\label{sec:finite-deployment}

For the softmax construction \eqref{eq:softmax}, the transfer result below extends the certificates of Sect.~\ref{sec:risk} from any parameter law, in particular a learned law of Sect.~\ref{sec:flow}, to one sampled network, up to its deployment error. We then check the requirement on the candidate directly by independent validation, whatever its actor family.

\subsection{From a learned law to one finite network}

Fix a parameter law $\mu$. Draw $X_1,\ldots,X_n$ independently from
$\mu$, independently of subsequent trajectories, and use the same
sampled units at every state and time, so that
\begin{equation}\label{eq:finite-score}
 f_n(y,a)=\frac1n\sum_{i=1}^n\psi(y,a,X_i),\qquad
 \pi_n(da\mid y)=
 \frac{e^{f_n(y,a)}\rho(da\mid y)}
      {\int e^{f_n(y,a')}\rho(da'\mid y)}.
\end{equation}
The action reference is unchanged, and only joint measurability and
$|\psi|\le L_0$ are needed.

\begin{theorem}[Population-to-network transfer]\label{thm:finite-deployment}
Let $\zeta\in(0,1)$ and
$\varepsilon_n=\min\{1,L_0/(2\sqrt{n(1-\beta)\zeta})\}$. With
probability at least $1-\zeta$ over the sampled units,
\begin{gather*}
 |q_\beta(\pi_n)-q_\beta(\pi_\mu)|\le\varepsilon_n,\qquad
 |V_m(\pi_n)-V_m(\mu)|\le\frac{m+\bar u}{1-\beta}\varepsilon_n,\\
 \max\{|V(\pi_n)-V(\pi_\mu)|,
        |U(\pi_n)-U(\pi_\mu)|\}
       \le\frac{u_-+\bar u}{1-\beta}\varepsilon_n
\end{gather*}
hold simultaneously. The last bound assumes $-u_-\le u\le\bar u$.
For any fixed $T$, a separate event of probability at least $1-\zeta$
gives
\begin{equation}\label{eq:finite-direct-horizon}
 |p_T(\pi_n)-p_T(\pi_\mu)|
 \le\min\left\{1,\frac{L_0\sqrt{T+1}}{2\sqrt{n\zeta}}\right\}.
\end{equation}
Using $\zeta/2$ in both estimates makes all conclusions simultaneous
with probability at least $1-\zeta$.
\end{theorem}

In the proof (Appendix~\ref{app:deployment-proofs}) we bound the expected one-step relative entropy from the population policy to the sampled network with the trajectory law fixed, so no covering of the state--action space is needed. This is a bound over independent draws of the units. Proposition~\ref{prop:finite-training} concerns continuous-time training under a common potential. Example~G.1 of Online Resource~1 shows the size of the bounds in a model in which the deviations are exact.

\subsection{Validate the deployed candidate at its specified horizon}
\label{sec:validation}

Condition on training and network construction, and draw $N$ fresh, independent trajectories of the fixed candidate, the sampled network \eqref{eq:finite-score} when a learned law and a sampler are available and otherwise the trained finite policy. The error probability $\zeta\in(0,1)$ of a test is chosen before validation and is distinct from the tolerance $\alpha$. When $K$ statements are to hold together, for example one for each of $K$ candidates or horizons, each test is run at the error probability $\zeta/K$. By this Bonferroni correction all $K$ statements hold simultaneously with probability at least $1-\zeta$. The number $K$ is fixed before a policy is selected. A failure to certify does not prove a violation. If the trajectories run through action $T$ and $j$ of them fail by $T$, the one-sided exact binomial upper bound is
\begin{equation}\label{eq:binomial-bound}
 \widehat\alpha=\sup\{p\in[0,1]:\Pprob(\mathrm{Bin}(N,p)\le j)\ge\zeta\},
 \qquad
 \Pprob(p_T\le\widehat\alpha)\ge1-\zeta .
\end{equation}
Even with zero observed failures, the upper bound is $1-\zeta^{1/N}$. A test at horizon $T$ certifies only the probability of failure by $T$.

Alternatively, let $\widehat V_H$ be the mean over the $N$ trajectories of the clipped return truncated after $H\ge1$ actions. Such a return lies in an interval of length $(m+\bar u)S_H$, with $S_H=(1-\beta^H)/(1-\beta)$, and the return after the first $H$ actions is at least $-m\beta^H/(1-\beta)$. By the one-sided Hoeffding inequality \citep{hoeffding1963}, $V_m\ge\widehat V_H-(m+\bar u)S_H\sqrt{\log(1/\zeta)/(2N)}-m\beta^H/(1-\beta)$ with probability at least $1-\zeta$, and substituting this lower bound in Eq.~\eqref{eq:risk-horizon-bound} gives the empirical certificate
\begin{equation}\label{eq:empirical-horizon-bound}
 \Pprob\left(p_T\le\min\left\{1,\
 \frac{\bar u-(1-\beta)\widehat V_H}{(m+\bar u)\beta^T}
 +\frac{1-\beta^H}{\beta^T}\sqrt{\frac{\log(1/\zeta)}{2N}}
 +\frac{m}{m+\bar u}\,\beta^{H-T}\right\}\right)\ge1-\zeta,
\end{equation}
whose three terms inside the minimum are the plug-in estimate, the sampling error and the truncation error. Since $\widehat V_H$ cannot exceed $\bar u(1-\beta^H)/(1-\beta)$, the first and last terms of Eq.~\eqref{eq:empirical-horizon-bound} sum to at least $\beta^{H-T}$, so the bound $p_T\le\alpha<1$ can be certified by the empirical certificate \eqref{eq:empirical-horizon-bound} only if
\begin{equation}\label{eq:validation-resolution-floor}
 \beta^{H-T}+
 \frac{1-\beta^H}{\beta^T}
       \sqrt{\frac{\log(1/\zeta)}{2N}}\le\alpha,
\end{equation}
which is checked before training. When the condition fails, one can draw more trajectories, follow them for longer or use the binomial test. The necessary condition is independent of $m$. Figure~G.2 in Appendix~G.1 of Online Resource~1 illustrates the resolution floor.

The binomial test and the empirical certificate can use the same $N$ trajectories when they include the required horizons. The binomial test uses only the indicator of failure by action $T$. It is exact, its bound at zero failures is of order $\log(1/\zeta)/N$, and it certifies the one horizon $T$. The wear pass counts use this construction with a Bernoulli event of failure by an independent geometric time. Each trajectory in that test has its own horizon draw. The empirical certificate uses the clipped returns to construct a lower confidence bound on the value in Proposition~\ref{prop:risk-value}. Its bounds hold for the geometric risk and for every horizon at once, and another valid lower bound on $V_m$ can take the place of the Hoeffding bound without new trajectories. The factor $\beta^{-T}$ amplifies its sampling error, which is of order $N^{-1/2}$. The binomial test is useful for a single horizon, particularly when few trajectories fail. Figure~G.2 of Online Resource~1 shows the terms of the bound \eqref{eq:empirical-horizon-bound} on Example~\ref{ex:running}.

For a task that physically ends after $H$ actions, with $-m$ charged
from failure until action $H-1$, $V_m^{(H)}$ the expected discounted
sum of the $H$ clipped rewards, and $T<H$,
\begin{equation}\label{eq:physical-horizon-risk}
 p_T\le\min\left\{1,
 \frac{\bar u S_H-V_m^{(H)}}
 {(m+\bar u)\beta^T(1-\beta^{H-T})/(1-\beta)}\right\}.
\end{equation}
The bound follows by taking expectations in the pathwise inequality $\sum_{t<H}\beta^t(\bar u-r_{m,t})\ge(m+\bar u)\beta^T(1-\beta^{H-T})/(1-\beta)$, which holds on $\{\tau\le T\}$ and whose left side is nonnegative on every path. In its empirical version, $V_m^{(H)}$ is replaced by a lower confidence bound from the $N$ complete returns. The Hoeffding sampling term is $(m+\bar u)S_H\sqrt{\log(1/\zeta)/(2N)}$ for the clipped reward. For a failure-only reward, the same argument uses $\max\{m,u_-\}$ in place of $m$ in the sampling term, since that reward pays utility in full before failure. The lander of Sect.~\ref{sec:lunar} uses a failure-only price at a physical horizon. A bound for its probability of failure by action $T$ can be obtained from the complete returns above or directly from fresh Bernoulli trials for $\{\tau\le T\}$.

\begin{proposition}[The risk priced by a physical horizon]\label{prop:physical-price}
For a task ending after $H$ actions, let $u_t$ be the reward before failure and $-m$ each reward from the failing action to action $H-1$, and define the truncated utility and the priced risk
\[
 U_H=\E\sum_{t<\min\{\tau,H\}}\beta^tu_t,\qquad
 q_{\beta,H}=\E\bigl[\beta^\tau(1-\beta^{H-\tau})\ind_{\{\tau<H\}}\bigr].
\]
\begin{enumerate}
\item[(i)] The truncated value decomposes as $V_m^{(H)}=U_H-\frac{m}{1-\beta}q_{\beta,H}$. Under the clipped-barrier reward with $m\ge u_-$, the interior loss $\mathcal I_{m,H}$, Eq.~\eqref{eq:interior-loss} truncated at $H$, is subtracted as well.
\item[(ii)] For $T<H$, $\beta^T(1-\beta^{H-T})p_T\le q_{\beta,H}$.
\item[(iii)] Parts (i)--(iii) of Proposition~\ref{prop:fixed-price} hold with $q_{\beta,H}$ and $U_H$ in place of $q_\beta$ and $U$, by the same argument, the second form of (ii) with the loose bound $U_H\le\bar u(1-q_{\beta,H})/(1-\beta)$.
\end{enumerate}
\end{proposition}

The proposition follows by direct computation. If the failure action is $\tau<H$, the price charged from failure to the horizon totals $m\beta^\tau(1-\beta^{H-\tau})/(1-\beta)$, and otherwise nothing is charged. Splitting the reward sum at $\tau$ and taking expectations gives~(i). On $\{\tau\le T\}$, $\beta^\tau-\beta^H\ge\beta^T-\beta^H$, which gives~(ii). The proof of Proposition~\ref{prop:fixed-price} uses only the identity \eqref{eq:exact-risk-scalarization}, so with the decomposition~(i) in its place it gives~(iii).

The risk $q_{\beta,H}$ weights a failure at $\tau$ by the discounted penalty periods that remain before $H$, so a failure shortly before the end costs little, and it increases to $q_\beta$ as $H\to\infty$. Since the price acts on $q_{\beta,H}$, the comparisons of (iii) do not assert that the exhaustion probability $p_{H-1}$ falls with $m$, because both the occurrence and the timing of failure are priced. A lump sum $F$ at failure instead prices $\E[\beta^\tau\ind_{\{\tau<H\}}]$. In Sect.~\ref{sec:lunar} the observations on the failure probability are read against this proposition.

\section{Experiments}\label{sec:numerics}

We report the findings as \emph{observations} (O1, O2, and so on). When an observation uses a theoretical result, we state its relevant conditions, including any bound on the optimization error. For each study we report every training seed and identify comparisons added after the initial runs. The settings were fixed before the runs, with the exception of the lander study (Sect.~\ref{sec:lunar}), where we report the trials used to choose the design. A policy passes if the one-sided exact binomial upper bound in Eq.~\eqref{eq:binomial-bound} is at most the tolerance. The Bonferroni correction in Sect.~\ref{sec:validation} controls the probability of any erroneous certification among the learned policies in a study at $0.05$. Online Resource~1 gives the test problems, experimental settings and further results. The replication repository contains the run data (Appendix~G.3 of Online Resource~1).

\subsectionbarrier
\subsection{An asset operated under a wear allowance}\label{sec:wear}

The model is that of Example~\ref{ex:running}. By dynamic programming on a grid we obtain policies for the interior barrier $B(z')=0.1\log(1+b/z')$ and for failure-only pricing at several values of $m$. We evaluate the interpolated policies in the continuous simulator. The resulting utility and risk of the failure-only policies trace a benchmark curve across prices. The learner is PPO \citep{ppo2017,sb32021} with fixed settings. A \emph{configuration} is a reward rule. It is a \emph{fixed price} $m$ with or without the interior barrier, where $m=0$ is training without a price, an \emph{adaptive multiplier}, or the \emph{action-regularized} configuration. The adaptive multiplier is a flat failure penalty. After every rollout it changes by the \emph{dual step}, a positive constant, times the difference between the failure frequency of the rollout and a tolerance $\alpha$, and it is kept nonnegative. The action-regularized configuration combines the low price $m=0.03$ with the action relative entropy, at the strength $\eta$, to the \emph{incumbent rule} $a=\min\{1,z/(3b)\}$ with a small Gaussian dispersion.

\suppressfloats[t]

\begin{table}[tb]
 \centering\footnotesize
 \caption{Learned policies in the wear-allowance example, eight seeds per configuration, with all probabilities in percent. The setting is the price $m$, the tolerance $\alpha$ of the adaptive multiplier, or the strength $\eta$ of the action regularization at the price $m=0.03$. Columns give $q_\beta$ and $U$ (mean $\pm$ between-seed standard deviation); bound est., a plug-in estimate of the right side of \eqref{eq:price-risk-law} with standing still as the comparison policy and simulated benchmark values, capped at one; the interior loss $\mathcal I_m$ under $B(z)=0.1\log(1+b/z)$; gap, the shortfall of $U$ from the interpolated benchmark curve at the estimated $q_\beta$ (mean and maximum); pass, the seeds whose binomial upper bound \eqref{eq:binomial-bound} on $q_\beta$ is at most $0.05$, with the Bonferroni correction over the $120$ learned policies; near, the frequency of coming within $0.05\,b$ of exhaustion; at $0.9\,b$, the breakdown probability when the true allowance is $0.9\,b$; and $\mathcal A$, the estimated action relative entropy to the incumbent rule before action clipping (last row).}
 \label{tab:wear-campaign}
 \setlength{\tabcolsep}{2.5pt}\begin{tabular}{@{}lrrrrrrrrr@{}}
\toprule
Setting & $q_\beta$ & bound est. & $U$ & $\mathcal I_m$ & gap & pass & near & at $0.9b$ & $\mathcal A$ \\
\midrule
\multicolumn{10}{@{}l}{Fixed price} \\
$m=0$ & $59.90\pm0.18$ & -- & $5.35\pm0.02$ & $1.01$ & $-0.2$ ($0.4$) & $0/8$ & $61.4$ & $62.8$ & -- \\
$m=0.1$\textsuperscript{$\dagger$} & $32.73\pm23.09$ & $100.0$ & $5.14\pm0.16$ & $1.63$ & $0.8$ ($2.9$) & $0/8$ & $49.5$ & $57.2$ & -- \\
$m=0.3$ & $4.42\pm0.75$ & $89.8$ & $4.76\pm0.05$ & $3.82$ & $0.9$ ($1.5$) & $3/8$ & $29.4$ & $45.1$ & -- \\
$m=1$ & $1.90\pm0.26$ & $27.2$ & $4.50\pm0.04$ & $3.68$ & $1.7$ ($2.9$) & $8/8$ & $12.6$ & $32.6$ & -- \\
$m=3$ & $0.97\pm0.24$ & $9.3$ & $4.22\pm0.08$ & $3.27$ & $3.8$ ($4.9$) & $8/8$ & $3.9$ & $17.7$ & -- \\
$m=10$ & $0.53\pm0.07$ & $3.0$ & $4.01\pm0.05$ & $3.06$ & $4.8$ ($6.3$) & $8/8$ & $2.0$ & $11.0$ & -- \\
\addlinespace
\multicolumn{10}{@{}l}{Fixed price with the interior barrier} \\
$m=1$ & $0.69\pm0.22$ & $33.7$ & $3.85\pm0.11$ & $2.66$ & $10.1$ ($13.3$) & $8/8$ & $1.0$ & $1.5$ & -- \\
$m=3$ & $0.37\pm0.08$ & $11.3$ & $3.70\pm0.10$ & $2.58$ & $9.7$ ($12.1$) & $8/8$ & $0.5$ & $0.9$ & -- \\
\addlinespace
\multicolumn{10}{@{}l}{Adaptive multiplier} \\
$\alpha=0.05$ & $5.03\pm0.44$ & -- & $4.78\pm0.03$ & $1.32$ & $1.3$ ($2.2$) & $0/8$ & $31.5$ & $46.2$ & -- \\
$\alpha=0.05$\textsuperscript{$\ddagger$} & $5.35\pm0.43$ & -- & $4.79\pm0.03$ & $1.33$ & $1.3$ ($2.1$) & $0/8$ & $34.2$ & $47.4$ & -- \\
$\alpha=0.01$ & $1.17\pm0.29$ & -- & $4.30\pm0.08$ & $1.39$ & $3.2$ ($4.4$) & $8/8$ & $5.5$ & $22.2$ & -- \\
$\alpha=0$ & $0.79\pm0.19$ & -- & $4.16\pm0.07$ & $1.39$ & $3.9$ ($5.1$) & $8/8$ & $3.2$ & $15.5$ & -- \\
\addlinespace
\multicolumn{10}{@{}l}{Action-regularized, $m=0.03$} \\
$\eta=0.01$\textsuperscript{\S} & $5.80\pm2.34$ & -- & $3.34\pm0.06$ & $1.80$ & $31.2$ ($33.0$) & $3/8$ & $10.7$ & $15.2$ & $29.9$ \\
$\eta=0.1$ & $1.47\pm0.16$ & -- & $2.80\pm0.02$ & $1.73$ & $37.9$ ($38.9$) & $8/8$ & $3.2$ & $5.7$ & $0.432$ \\
$\eta=1$ & $1.13\pm0.09$ & -- & $2.73\pm0.01$ & $1.72$ & $38.4$ ($39.4$) & $8/8$ & $2.6$ & $4.7$ & $0.00709$ \\
\midrule
Incumbent rule & $1.08$ & -- & $2.73$ & -- & -- & -- & -- & -- & $0$ \\
\bottomrule
\multicolumn{10}{@{}p{\linewidth}@{}}{\textsuperscript{$\dagger$}three seeds stayed near the risk of the no-price policy ($q_\beta=0.59$), four reached $0.11$ to $0.17$ and one $0.33$; the mean is of a two-cluster sample. \textsuperscript{$\ddagger$}dual step $1$ instead of $5$. \textsuperscript{\S}the breakdown probability differs between the seeds, from $2.9$ to $8.6$ percent, and was still falling when training ended; the grid benchmark for the action-regularized objective has $1.36$ percent.} \\
\end{tabular}

\end{table}

We show the runs in Fig.~\ref{fig:wear-frontier} and Table~\ref{tab:wear-campaign}. We make the following observations.
\begin{itemize}[leftmargin=2.8em,itemsep=0pt]
\item[(O1)] The plug-in estimates of the right side of Eq.~\eqref{eq:price-risk-law} exceed the estimated risks in the fixed-price and interior-barrier runs. They use benchmark values in place of the continuous-model supremum and objective gap. The pass column uses independent binomial validation.
\item[(O2)] Across the failure-only prices, the mean estimated risk falls and the mean stopped utility does not rise. Proposition~\ref{prop:fixed-price}(iii) gives monotonic risk for exact failure-only optimizers. For learned policies it allows an increase proportional to their objective gaps, so the decrease reported here is an observation.
\item[(O3)] The fixed-price policies without the interior barrier and the adaptive-multiplier policies have stopped utility close to the interpolated benchmark curve at their estimated risks. Table~D.2 of Online Resource~1 reports the separate comparison at each policy's own price.
\item[(O4)] The interior-barrier policies have lower stopped utility than the failure-only benchmark at comparable estimated risks. Theorem~\ref{thm:matched-risk} identifies the comparison policy's interior loss as one term in a utility comparison under its stated hypotheses.
\item[(O5)] At the tested prices, the interior-barrier policies came close to exhaustion less often at the nominal allowance and broke down less often at a reduced allowance than the failure-only policies. For the tested reduction, the failure-only benchmark computed using the reduced allowance gives a similar reduction in breakdowns at a smaller utility cost. Its protection is weaker when the actual reduction is larger (Fig.~D.2 of Online Resource~1).
\item[(O6)] Proposition~\ref{prop:action-risk-bridge} bounds the deviation from the reference risk in terms of the achieved action relative entropy. In the runs at the two larger action strengths, the estimated risks are near the incumbent's and the estimated relative entropies give informative bounds. The stopped utilities lie below the benchmark curve at these risks.
\end{itemize}
The estimated risks of the adaptive-multiplier policies are near their prescribed tolerances. For the configurations trained at tolerance $0.05$, none of the seeds has a binomial upper bound below that tolerance. At tolerance zero the multiplier increased in every seed. Positive throughput exposes the asset to an unbounded wear shock.

At prices $0$, $0.3$ and $1$, the breakdown frequency leveled off and the objective was close to the grid benchmark. The interior-barrier runs also had objective estimates close to their benchmark. At the price $0.1$ and at the weakest action regularization, some traces were still changing when training ended. The higher-price failure-only runs leveled off at higher estimated breakdown risks than their corresponding benchmarks. Repetitions under a constant learning rate and own-price comparisons are reported in Appendix~D.2 of Online Resource~1.

\subsectionbarrier
\subsection{A learned lander paid for each landing}\label{sec:lunar}

In the second study a learned controller lands a vehicle with a limited supply of fuel. We define the landing problem using the dynamics of LunarLanderContinuous-v3 from Gymnasium, with a new reward and failure event. A successful landing on the pad yields revenue of $36$, and each unit of fuel costs $1$. The tank holds $24$ units. A flight fails if the lander crashes, leaves the frame or uses more than its fuel budget before the first leg touches the ground. A failed flight yields no landing revenue. In the model of Sect.~\ref{sec:model}, the loss of the lander is represented by the exhaustion of a second budget. The landing revenue is about twice the fuel cost of a typical landing, and the tank holds roughly one-third more fuel than a landing uses.

We use a failure-only price. The reward before failure is the full utility, and the price is charged from the failing frame to the physical horizon. Proposition~\ref{prop:physical-price} gives the corresponding priced risk $q_{\beta,H}$. The prices through charge $4$ are below the largest one-frame fuel cost, so preserving utility before failure is material in this example. We index prices by the \emph{charge} $k$, the charge of a failure at frame $100$ in units of the revenue, $k=m(1-\beta^{900})/(36(1-\beta))$. The charges are $k\in\{0,0.25,0.5,1,2,4,8,16\}$, with ten seeds each.

We train the lander policy with PPO and normalize the rewards \citep{ppo2017,sb32021}. During training the reward also contains the change of the simulator's potential for the approach to the pad, scaled down and set to zero at the end of an episode. Its discounted sum over an episode is minus its value at the start, on every path, so it changes the value of every policy by the same amount and leaves the problem unchanged. At every charge we continue training from the policy of the same seed trained without a price. The charge therefore acts on a policy that already lands successfully. The learning rate is constant and is decayed to zero in a final phase. Before the decay, fuel use during training is roughly stable through charge $4$. At the two largest charges, fuel use continues to rise and utility falls slowly (Appendix~D.3 of Online Resource~1). Our results hold for every learner through its optimization error, so the learner of a study can be chosen to suit its task.

\begin{figure}[tb]
 \centering
 \includegraphics[width=\linewidth]{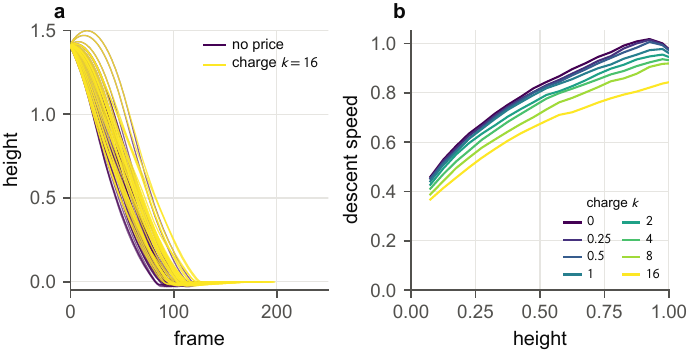}
 \caption{The landing problem. (a)~The height against the frame in the $32$ flights of one seed without a price and at the charge $k=16$, in the coordinates of the simulator's observation, with the pad at height zero. All $64$ flights land. The seed is the one whose failure probability, summed over the two charges, is the median of the ten seeds. (b)~The descent speed against the height, the median over the frames of $320$ flights at each charge, $32$ per seed. Figure~D.6 of Online Resource~1 shows each seed.}
 \label{fig:lander-flights}
\end{figure}

\begin{figure}[tb]
 \centering
 \includegraphics[width=\linewidth]{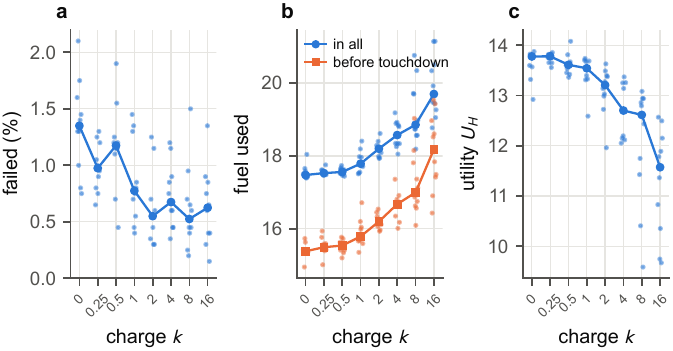}
 \caption{The landing problem, ten seeds per charge $k$ and $2000$ evaluation episodes per policy; small markers are the seeds and lines their medians. (a)~Failure probability. (b)~Total fuel use per episode and fuel use before the first touchdown. (c)~Stopped utility $U_H$. The values are given in Table~D.3 of Online Resource~1.}
 \label{fig:lander-summary}
\end{figure}

We show the flights of one seed in Fig.~\ref{fig:lander-flights} and all runs in Fig.~\ref{fig:lander-summary}. We make the following observations.
\begin{itemize}[leftmargin=2.8em,itemsep=0pt]
\item[(O7)] At higher tested charges the lander descends more slowly and uses more fuel before touchdown. Mean failure frequency is lower at charge $2$ than without a price and changes little across higher tested charges. Under exact failure-only optimization, the priced risk $q_{\beta,H}$ and stopped utility are nonincreasing by the two optimality comparisons behind Proposition~\ref{prop:physical-price}(iii). Failure frequency and fuel use are separate measurements. The estimated priced risk rises at three adjacent charge steps, and estimated stopped utility rises at one (Table~D.3 of Online Resource~1). Eq.~\eqref{eq:price-monotone} allows increases in priced risk for approximate optimizers.
\item[(O8)] Among the evaluated policies, those trained at charge $0.25$ have the highest estimated mean value at evaluated charges through $4$, and those trained at $2$ have the highest at $8$ and $16$. The shortfall from the best tested value at each policy's own charge grows from charge $0.25$ onward (Table~D.4 of Online Resource~1). This comparison estimates a lower bound on the gap to the unknown optimum. It leaves the trend in the true gaps and the cause of the extra fuel use unresolved. The comparison runs used a shorter training budget and started from a policy trained on the simulator's original task. At every tested charge, their median value evaluated at that charge was lower (Table~D.5 of Online Resource~1).
\end{itemize}

\subsectionbarrier
\subsection{The convergence theorem and the regularizers on the analytical example}
\label{sec:flow-test}

\begin{figure}[tb]
 \centering
 \includegraphics[width=0.9\linewidth]{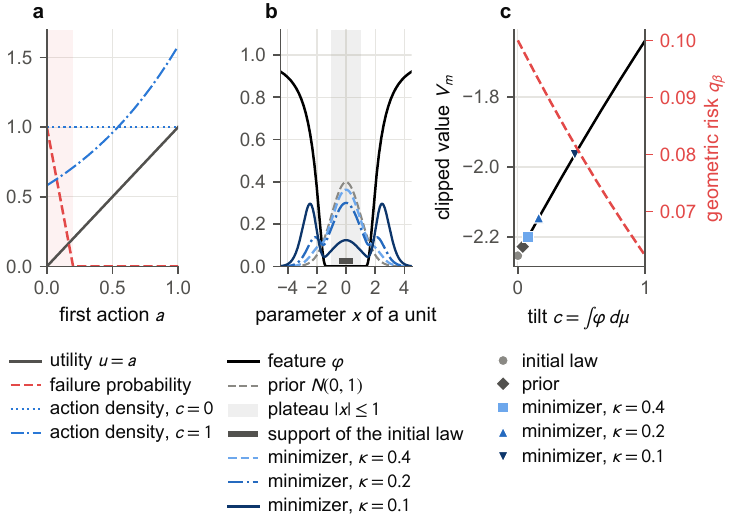}
 \caption{Setup of the analytical example ($\varepsilon=0.2$, $\beta=0.9$, $m=1$). (a)~Against the first action $a$, the utility, the failure probability (positive in the shaded range $a<\varepsilon$) and the action densities of the tilts $c=0$ and $c=1$, the two ends of the represented policies. (b)~The feature $\varphi$ of a unit, the density of the prior, the support of the initial law, and the densities of the minimizers $\mu^*$ under relative entropy at three strengths $\kappa$. The feature vanishes on the plateau $|x|\le1$ (shaded). (c)~The clipped value $V_m$ (solid, left axis) and the geometric risk $q_\beta$ (dashed, right axis) against the tilt $c$, with the tilts of the initial law, the prior and the three minimizers marked.}
 \label{fig:example-setup}
\end{figure}

Theorem~\ref{thm:flow} concerns the gradient flow of the regularized objective, so we study it on a one-dimensional problem with a closed-form action density for each tilt. We characterize the relative-entropy minimizer by a scalar problem and approximate the population flow by a finite-volume scheme. We call the problem the \emph{analytical example} and show its setup in Fig.~\ref{fig:example-setup}. The example illustrates the theorem and the roles of the prior and of the strength. It is not a model of an application.

The state has a phase, which is zero at the start and one after the first action, and a draw $s_2$, which is uniform on $[0,1]$. The action set is $[0,1]$. In phase zero an action $a$ earns the utility $a$ and consumes $1-a+\varepsilon s_2$ of the budget $b=1$, for a constant $\varepsilon\in(0,1/2)$. In phase one nothing is earned or consumed. With the cost discount $\gamma=1/2$ in Eq.~\eqref{eq:budget}, the budget is exhausted exactly when $a<\varepsilon s_2$. Only the first action therefore has an effect. It fails with probability $\max\{0,1-a/\varepsilon\}$, so a larger action earns more and fails less often (Fig.~\ref{fig:example-setup}(a)).

The barrier is $B(z)=\log(1+1/z)$, and the action reference is uniform on $[0,1]$. A unit with parameter $x\in\R$ has the score $a\varphi(x)$ with the feature $\varphi(x)=\exp[-(|x|-1)^{-2}]$ for $|x|>1$ and $\varphi(x)=0$ on the \emph{plateau} $|x|\le1$ (Fig.~\ref{fig:example-setup}(b)). By Eq.~\eqref{eq:softmax}, the policy of a parameter law $\mu$ is the same in every state. It is the exponential tilt of the uniform reference,
\[
 \pi_c(da)=\frac{e^{ca}\,da}{\int_0^1e^{ca'}\,da'},
 \qquad c=c(\mu)=\int\varphi\,d\mu\in[0,1).
\]
The learner therefore controls one number, the tilt $c$. As $c$ grows, the clipped value rises and the risk falls (Fig.~\ref{fig:example-setup}(c)). Since $c<1$, the supremum of the clipped value is not attained. A unit on the plateau has no effect on the policy and receives no value gradient.

The prior is $\Gamma=N(0,1)$, the parameter regularizer is $\Hc_{\omega,G}$ with the strength $\kappa$, and there is no action regularization. We use $\varepsilon=0.2$, $\beta=0.9$ and $m=1$. Without regularization every unit follows its value gradient, which we call \emph{unregularized transport}. A unit on the plateau then does not move. Our runs start from the prior conditioned on $[-1/2,1/2]$, a law on the plateau, so that the value gradient gives no help at the start. Under relative entropy alone the minimizer $\mu^*$ of the objective \eqref{eq:training-objective} is an exponential tilt of the prior by the feature, and it is computed from a problem in one variable. A larger strength gives a minimizer closer to the prior, and a smaller strength moves mass off the plateau and raises the tilt (Fig.~\ref{fig:example-setup}(b) and~(c)). For the divergence regularizers used here, the population equation is a one-dimensional drift--diffusion equation. The \emph{observed rate} of a run is the least-squares slope of the logarithm of a decaying quantity, and the \emph{objective gap} is the excess of the objective \eqref{eq:training-objective} over its minimum. When the particle system \eqref{eq:particle} with relative entropy is simulated on this gradient flow, its deviation from the finite-volume solution shrinks with the width (Appendix~F.2 of Online Resource~1). Appendices~F and~D.4 of Online Resource~1 give the construction in full and verify the properties stated here.

\suppressfloats[t]
\begin{figure}[tb]
 \centering
 \includegraphics[width=0.9\linewidth]{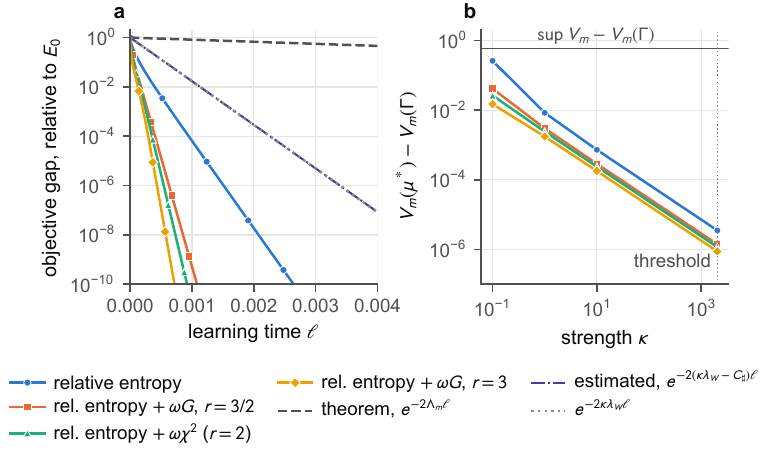}
 \caption{The gradient flow of the analytical example for relative entropy alone and for $\Hc_{\omega,G}$ with the power integrands $r\in\{3/2,2,3\}$, of which $r=2$ is the $\chi^2$ divergence of Eq.~\eqref{eq:flow-divergence} (prior $N(0,1)$, $m=1$, $\omega=1$). (a)~The objective gap $\Jc_m(\mu_\ell)-\Jc_m(\mu^*)$ between the gradient flow's law $\mu_\ell$ and the minimizer $\mu^*$, relative to its initial value $E_0$ of Eq.~\eqref{eq:flow-energy-rate}, at the strength $\kappa=1.05\,C_m/\lambda_W$, five percent above the threshold. The dashed line is the envelope $e^{-2\Lambda_m\ell}$ of Eq.~\eqref{eq:flow-energy-rate}, the dash-dotted line the envelope with a curvature estimate of the example's own value curve (Appendix~D.4 of Online Resource~1), and the dotted line $e^{-2\kappa\lambda_W\ell}$. (b)~Improvement of the minimizer over the prior, $V_m(\mu^*)-V_m(\Gamma)$, against $\kappa$; the horizontal line is the largest possible improvement. Panel (a) supports (O9) and panel (b) supports (O10).}
 \label{fig:regularizer-flow}
\end{figure}

\begin{figure}[tb]
 \centering
 \includegraphics[width=0.92\linewidth]{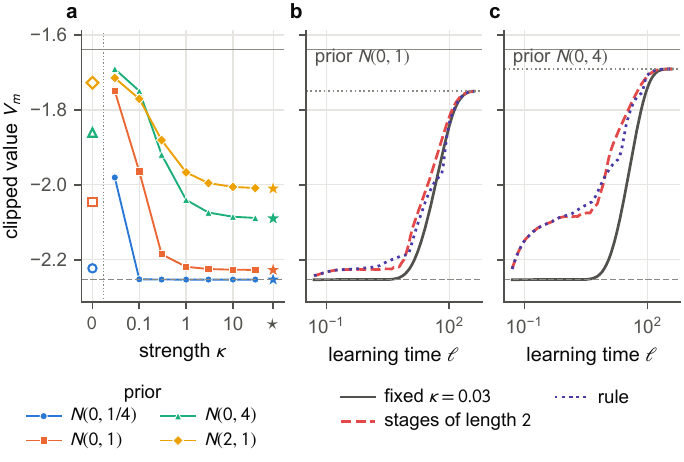}
 \caption{The prior, the strength and a strength lowered in steps in the analytical example (relative entropy alone, $m=1$). (a)~Clipped value of the minimizer against the strength $\kappa$ for four priors. The stars at the right end are the prior policies, the limit of large $\kappa$. The open markers at the left end are the limits of unregularized transport started from each prior. (b),~(c)~Clipped value against the learning time under the priors $N(0,1)$ and $N(0,4)$, from the initial law on the plateau, for the strength $0.03$ held fixed and for a strength lowered through $10$, $3$, $1$, $0.3$, $0.1$ and $0.03$ (means over eight seeds of $10{,}000$ units). A stage is the learning time at one strength. We use either stages of length $2$ or a rule that ends a stage when the mean clipped values in two consecutive windows of length $0.5/\kappa$ differ by less than $0.002$. In all panels the solid horizontal line is the supremum of the clipped value and the dashed line is the value of the initial law, and the dotted lines of (b) and (c) are the clipped values of the minimizers at $\kappa=0.03$. Panel (a) supports (O11), and panels (b) and (c) support (O12).}
 \label{fig:prior-strength}
\end{figure}

We show the runs in Figs.~\ref{fig:regularizer-flow} and~\ref{fig:prior-strength}. In (O9) and (O10), the observed rates and computed minimizers illustrate the bounds of Theorem~\ref{thm:flow} and the analytical example. The runs of (O11) lie below the sufficient threshold.
\begin{itemize}[leftmargin=2.8em,itemsep=0pt]
\item[(O9)] Above the threshold the objective gap decays at least at the guaranteed rate $2\Lambda_m$, for relative entropy and the power integrands of Proposition~\ref{prop:generators} alike. The observed rates exceed the guaranteed rate by orders of magnitude and vary across integrands. The bound uses $C_m$, a uniform upper bound on the curvature deficit of the negative clipped value.
\item[(O10)] The guaranteed rate and threshold are common to all integrands. The minimizer and the regularization bias depend on the integrand. With an integrand of faster growth the minimizer stays closer to the prior.
\item[(O11)] For a Gaussian prior $N(x_0,\sigma^2)$ the observed rates nearly coincide at equal $\kappa/\sigma^2$, the weight-decay strength. The guaranteed rate depends on $(\kappa,\sigma)$ only through this ratio (Sect.~\ref{subsec:prior}), whereas the minimizer depends on both and tracks the prior policy (Fig.~D.8 of Online Resource~1).
\end{itemize}
Below the threshold the computed objective gaps also decrease on the tested runs, and the risk bounds exceed the measured deviations from the prior (Appendix~D.4 of Online Resource~1).

In these runs the role of the prior and the role of the strength can be distinguished (Fig.~\ref{fig:prior-strength}(a)). As the strength decreases, the clipped value of the minimizer rises from that of the prior policy toward the supremum of the clipped value, as Proposition~\ref{prop:decreasing-strength} states for the limit. At the largest tested strengths the computed clipped value is close to the prior's. Smaller strengths yield higher clipped values in these comparisons, and the fitted rates fall as the weight-decay strength decreases. The prior needs mass where the feature is not flat. Under a prior concentrated on the plateau the minimizer stays near the uniform policy at every strength of the runs except the smallest. Under relative entropy every unit also receives noise, with which the units on the plateau move, as they never do in unregularized transport. The strength can also be lowered in steps during training, so that a run trains first at larger strengths and then at the smallest tested strength (Fig.~\ref{fig:prior-strength}(b) and~(c)).

\begin{itemize}[leftmargin=2.8em,itemsep=0pt]
\item[(O12)] For the two priors in panels (b) and (c), the scheduled runs reach clipped values close to the computed value of the minimizer at the smallest tested strength. The measured times are generally shorter than under the fixed final strength and the gain is larger for the wider prior. The rule's gain for the narrower of these two priors lies within the between-seed spread. This observation was formulated after the runs.
\end{itemize}
These are finite particle simulations. Proposition~\ref{prop:decreasing-strength} concerns near-minimizers with vanishing objective gaps. Theorem~\ref{thm:flow} gives a rate within each exact-flow stage of positive net curvature.

In the analytical example, the quadratic potential alone moves the units toward its center and leaves the clipped value at the plateau value. Under diffusion the units can reach the region where the feature varies. Appendix~D.4 of Online Resource~1 reports the particle simulations. We also train policies for the wear-allowance problem with PPO and a quadratic anchor. We fit the reference parameters to the incumbent rule. This finite-network implementation applies the anchor to each actor parameter and falls outside the hypotheses of Proposition~\ref{prop:finite-training}. The risk and utility traces at fixed positive strengths approach different levels. At larger strengths those levels are closer to the incumbent's. When only the mean network is anchored, the fitted dispersion leads to nearly bang-bang control in the configuration tested. The effect of a given strength depends on the optimizer and on how the penalty enters the update.

With persistent noise, the performance of individual policies varies during training. Under the schedules, the stages with stronger anchors end close to the corresponding fixed-strength results. Some of the weaker stages still fall short of those results. With stochastic gradient descent, the run without an anchor reaches the comparison utility sooner than those schedules. Appendices~D.5 and~D.6 of Online Resource~1 give the training settings and further results.
\section{Choice of method, scope and conclusion}\label{sec:guidance}

\subsection{Which formulation to prefer}\label{sec:prefer}

A fixed price and an explicit constraint can lead to the same policy. To compare them, we use the same policy class, utility and cost, with the same convention for stopping at failure. A maximizer of the priced objective also solves the constrained problem if it is feasible and satisfies complementary slackness. A constrained optimum can be recovered by pricing when it has a finite supporting multiplier. If several policies maximize the priced objective at that multiplier, feasibility and complementary slackness identify which of them also solve the constrained problem. A restricted policy class can have constrained optima with no supporting price.

Suppose the dynamics and policies are deterministic and the initial state is fixed. Each policy then produces a single consumption total. At the same budget and horizon, a cap on expected consumption, a CVaR cap and a chance constraint with tolerance below one have the same feasible policies. Charging linearly for that total yields the corresponding fixed-multiplier Lagrangian. A running charge on the remaining reserve can assign greater weight to early consumption, and clipping limits the charge when the reward reaches its floor. Under stochastic consumption, equal means can conceal differences in the frequency and size of budget exceedances. Even at the same probability of failure by $T$, policies can have different geometric risks if their failure times differ.

CPO addresses an expected-cost constraint using a local model at each policy update. The cost can be chosen to match the discounted absorbing failure cost in our objective. In that case, a trust-region step at a fixed price can coincide with the CPO proposal when the local gradients, metric and effective multiplier agree. The cost slack can change after the update, so the agreement need only hold at the current iterate. Appendix~E.1 of Online Resource~1 gives the cost definitions and the local step.

We use a fixed price when the failure charge represents an operational cost, its time profile is appropriate for the task, and exploratory failures are permissible in training. A compatible learner can then use the bounded reward on the augmented state. An interior barrier is appropriate when retaining a reserve has value, for example when the actual allowance may be smaller than anticipated. Under failure-only pricing, pre-failure utility is unchanged. A lump sum $F$ at failure is equivalent to the continuing per-period charge $m=(1-\beta)F$. At a finite physical horizon the equivalent lump sum depends on the failure time.

CPO can also be used with an interior barrier. If low reserves should reduce utility and failure risk is subject to a prescribed limit, we use stopped utility net of interior loss as the objective and the prescribed risk as the constraint. The CPO update then determines a local cost multiplier. If failure also has an operational cost, the fixed charge can be retained in the objective under the cap. This changes the objective and can leave the cap slack. The resulting policy can be validated at the deployment horizon independently of the learning algorithm.

When a limit is specified in advance, it determines which cost to constrain. The optimization algorithm is a separate choice. CPO is appropriate when approximate satisfaction of an expected-cost limit is required at intermediate updates and the local constraint and trust region can be estimated. An adaptive multiplier can be used when there is feasibility slack and the dual step can be calibrated. For a deployment-horizon tolerance, the relevant event is failure by that horizon. It can be used as the constrained cost or evaluated independently for the returned candidate. A CVaR constraint limits the average severity of the worst outcomes. A learning algorithm for this formulation must estimate or otherwise account for the tail of the cost distribution. If individual failures are impermissible, pathwise feasibility must be established, for example with a known controlled-invariant set. When the augmented model is known and tractable, one can solve the chosen formulation directly.

To compare policies trained under different formulations, we evaluate all of them using the same deployment utility and risk criterion. Under uncertain consumption, for example, a comparison of expected consumption leaves failure probabilities undetermined. A value-based certificate requires a valid bound on value. The price--risk comparison also requires a justified bound on the objective gap. Under exact failure-only optimization with unclipped pre-failure utility, geometric risk is nonincreasing in the price. Proposition~\ref{prop:fixed-price}(iii) bounds any increase under an interior barrier or inexact optimization.

When the tolerance is uncertain, training at several prices and validating the candidates helps assess the trade-off. The clipped value remains bounded even when no policy meets the tolerance. In Sect.~\ref{sec:wear} the multiplier at tolerance zero grew throughout training. Eq.~\eqref{eq:price-gap-at-stake} bounds the loss from using a lower-price policy at a higher price. In the lander runs the higher tested charges gave similar failure rates with greater fuel use and larger shortfalls from the best tested values. Scaling and training accuracy should be assessed at the prices under consideration. A policy trained without a price is also a candidate if it meets the tolerance on independent data.

\subsection{Fixed strengths and schedules}\label{sec:schedules}

At fixed regularization strengths, the population gradient flow of Theorem~\ref{thm:flow} has a nonincreasing regularized objective. Above its sufficient curvature threshold, it converges exponentially to the unique minimizer of that objective. The clipped value of the minimizer falls short of that of any comparison law by at most the regularization cost of the comparison law, as follows by comparing their objectives in Eq.~\eqref{eq:training-objective}. A larger parameter strength puts more weight on the parameter cost in this bound and can improve the guaranteed rate.

Proposition~\ref{prop:decreasing-strength} concerns near-minimizers at a fixed price and a fixed action regularization weight. If the parameter strengths and optimization errors both tend to zero under its conditions, the objective values with the parameter term removed approach the infimum of that objective. Each cluster point minimizes it. If the action weight is zero, the limit is the supremum of the clipped value, which need not be attained. If the strength is lowered in steps and each stage follows the exact flow, the current objective remains nonincreasing. Within any stage above the curvature threshold, the exponential bound applies at that stage's fixed strengths. The result on near-minimizers applies to the stage-end laws if the strengths and their objective gaps vanish. This result requires control of the objective gap at the end of each stage.

In the analytical example, scheduled runs reached the final clipped-value target in less time on average for two tested priors (O12). For a third prior, none of the configurations reached that target within the run. In the wear-allowance study, PPO with stochastic gradient descent reached the comparison utility sooner without the anchor than with a schedule that reduced it. The stages nevertheless gave policies with different risks and utilities. These observations suggest keeping each candidate, monitoring utility and risk across stages, and comparing a schedule with training at the final strength from the start. The two studies used different regularizers and learning dynamics. Appendix~G.2 of Online Resource~1 gives the comparisons and discusses the choice of stage lengths.

A positive final strength can be useful when a trusted prior or better conditioning is worth its effect on the objective. If the price and action weight specify the desired objective, removing the parameter term gives that objective for training. A finite run at zero strength still requires evaluation of the returned policy. Candidate selection can be based on validation at the deployment horizon, with the error probability allocated as in Sect.~\ref{sec:validation}.

\subsection{Scope of the guarantees}\label{sec:scope}

The value certificate of Proposition~\ref{prop:risk-value} needs a valid value bound, and direct validation needs fresh trajectories of the candidate. Both can be used regardless of the learner. Theorem~\ref{thm:matched-risk} and Proposition~\ref{prop:fixed-price} additionally need a justified objective comparison. Theorem~\ref{thm:finite-deployment} bounds changes in risk and value for one network constructed by sampling independent units from a softmax parameter law. Theorem~\ref{thm:flow} concerns the exact population gradient flow. Under a common smooth potential, width-scaled continuous-time training follows the same gradient flow from the network's empirical law (Proposition~\ref{prop:finite-training}). When $\Lambda_m>0$, the limit is a point mass and all units converge to the same parameter. The finite-width limit for entropy-only Langevin training remains open here. Section~\ref{sec:flow-test} reports decreasing deviations at the tested widths. A guarantee that a noisy finite network with discrete steps and PPO's or SAC's estimators reaches the regularized optimum of the clipped value remains open. The threshold of Theorem~\ref{thm:flow} is sufficient and conservative, and in the analytical example the minimizer at the threshold lies close to the prior. The sampling width also far exceeds the widths used experimentally (with $L_0=1$, $\beta=0.95$ and $\zeta=0.05$, a transfer error of one percent requires about a million units). For deployment we therefore rely on direct validation. Joint adaptation of price, reference and regularization has no convergence guarantee here. Every certificate and validation is conditional on the simulator. We examine model error only in the runs of Sect.~\ref{sec:wear} in which the true allowance is smaller than the allowance used in training (O5), and validation across model parameters is the natural extension.

\subsection{Conclusion}

A clipped barrier reward puts a price on geometric failure risk. A maximizer of the clipped value maximizes utility net of interior loss among the policies of no greater risk. The candidate's risk can be assessed from a valid value bound or independent validation (Propositions~\ref{prop:risk-value} and~\ref{prop:fixed-price}, Theorem~\ref{thm:matched-risk}).

Our main convergence result applies to parameter regularizers with sufficient curvature along generalized geodesics, including relative entropy, its displacement-convex extensions and convex potentials (Theorem~\ref{thm:flow}, Proposition~\ref{prop:generators}). The rate and threshold depend on the regularizer through its curvature constant. Compared with the relative-entropy setting of \citet{leahy2022}, the general formulation lets us distinguish the effects of the parameter potential and of diffusion in the particle system. For the action policy, the achieved relative entropy to a reference bounds the difference in geometric failure risk. Starting the population gradient flow at the prior gives a further risk bound under the hypotheses of Proposition~\ref{prop:online-baseline}.

We bound the changes in value and risk when the units of a finite network are sampled independently (Theorem~\ref{thm:finite-deployment}). For a fixed deployed policy, we assess failure risk on fresh trajectories under a separate allocation of error probability. In the experiments we compare the bounds with the grid benchmark and the analytical example, and apply the procedure to learned policies. A practitioner can keep a compatible learner in use, add the reward wrapper and validate the resulting policy at the deployment horizon. Sharper curvature bounds and the analysis of noisy finite networks are directions for further work.

\appendix
\numberwithin{equation}{section}
\numberwithin{figure}{section}
\numberwithin{table}{section}
\section{Proofs of the risk certificates and performance comparisons}
\label{app:performance-proofs}

All policy comparisons use the same stochastic model. The proofs allow random transitions and an action reference with positive failure risk.

\subsection{The identity and the risk certificates}

\begin{proof}[Proof of Eq.~\eqref{eq:exact-risk-scalarization}]
Because costs are nonnegative, a negative reserve persists, so $r_m=-m$ at and
after $\tau$. Since $u+m\ge0$, before failure
$r_m=u-\min\{B(z_{t+1}),u+m\}$. Splitting the discounted series at $\tau$ gives
\[
 \sum_{t\ge0}\beta^tr_m(y_t,a_t)
 =\sum_{t<\tau}\beta^tu(s_t,a_t)-\sum_{t<\tau}\beta^t\min\{B(z_{t+1}),u+m\}
 -m\sum_{t\ge\tau}\beta^t,
\]
On $\{\tau<\infty\}$, the last series is $\sum_{t\ge\tau}\beta^t=\beta^\tau/(1-\beta)$. Taking expectations of the three terms gives $U$, $\mathcal I_m$ and $mq_\beta/(1-\beta)$. This proves Eq.~\eqref{eq:exact-risk-scalarization}. The signed series are
absolutely integrable, since $|u|\le\max\{u_-,\bar u\}$ and
$0\le\min\{B,u+m\}\le m+\bar u$. By the latter bound, Tonelli's theorem also applies to
the nonnegative interior loss.
\end{proof}

\begin{proof}[Proof of Proposition~\ref{prop:risk-value}]
Since $r_m=-m$ at and after $\tau$ and $0\le\bar u-r_m\le m+\bar u$,
\[
 \sum_{t\ge0}\beta^t(\bar u-r_m(y_t,a_t))
 \ge\frac{m+\bar u}{1-\beta}
             \beta^\tau\ind_{\{\tau<\infty\}}.
\]
The map taking a time $t$ and a trajectory to
$\beta^t(\bar u-r_m(y_t,a_t))$ is nonnegative and measurable for the
product of counting measure in time and the trajectory probability measure.
By Tonelli's theorem, expectation and the series can therefore be exchanged. Both are
finite, bounded by $(m+\bar u)/(1-\beta)$. The signed reward series is
absolutely integrable because $|r_m|\le\max\{m,\bar u\}$. Consequently
\[
 \frac{\bar u}{1-\beta}-V_m(\pi)
 \ge\frac{m+\bar u}{1-\beta}q_\beta(\pi).
\]
Also $\beta^\tau\ind_{\{\tau<\infty\}}\ge
\beta^T\ind_{\{\tau\le T\}}$, so $p_T\le\beta^{-T}q_\beta$.
This proves the two certificates. For a represented law,
nonnegativity of $\mathcal R$ and Eq.~\eqref{eq:direct-certificate-step0} give
$V_m(\widehat\mu)\ge-\widetilde\Jc_m(\widehat\mu)
-\varepsilon_r/(1-\beta)$, proving Eq.~\eqref{eq:objective-certificate}.
Finally, for an independent $\mathcal T$ with $\Pprob(\mathcal T\ge t)=\beta^t$,
independence and countable additivity give
$\Pprob(\tau\le\mathcal T)=\sum_{t\ge0}\Pprob(\tau=t)\beta^t=q_\beta$, the form of the geometric risk used in Sect.~\ref{sec:risk}.
\end{proof}

\begin{proof}[Proof of the failure-time bound of Remark~\ref{rem:failure-time}]
Let $\theta=\log(1/\beta)$. If $q_\beta=0$, then $\tau=\infty$ almost surely and the lower bound is $+\infty$. Otherwise the mean failure time over the earliest $\varepsilon$-fraction of trajectories is the infimum of $\E[\zeta\tau]$ over densities $\zeta\ge0$ with $\E\zeta=1$ and $\zeta\le1/\varepsilon$, the dual form of the lower tail average. Fix such a $\zeta$ and $n\ge1$. Jensen's inequality for the convex function $t\mapsto e^{-\theta t}$ under the probability $\zeta\,d\Pprob$ gives
\[
 e^{-\theta\E[\zeta(\tau\wedge n)]}\le\E[\zeta e^{-\theta(\tau\wedge n)}]\le\varepsilon^{-1}\E e^{-\theta(\tau\wedge n)},
\]
and $\E e^{-\theta(\tau\wedge n)}=\E[\beta^\tau\ind_{\{\tau\le n\}}]+\beta^n\Pprob(\tau>n)\to q_\beta$. Monotone convergence in $n$ gives $\E[\zeta\tau]\ge\log(\varepsilon/q_\beta)/\theta$ for every admissible $\zeta$, which is the claim. At $\varepsilon=p_T>0$ the density $\zeta=\ind_{\{\tau\le T\}}/p_T$ is admissible and gives a mean of at most $T$, so $T\ge\log(p_T/q_\beta)/\theta$, which is $p_T\le\beta^{-T}q_\beta$.
\end{proof}

\subsection{The matched-risk comparison}

\begin{proof}[Proof of Theorem~\ref{thm:matched-risk}]
Retaining both regularization costs in the objective comparison gives
\[
 V_m(\bar\mu)-V_m(\widehat\mu)
 \le\mathcal R(\bar\mu)-\mathcal R(\widehat\mu)
                 +\delta+\frac{2\varepsilon_r}{1-\beta}.
\]
Substituting \eqref{eq:exact-risk-scalarization} for both policies gives
\[
 U(\pi_{\bar\mu})-U(\pi_{\widehat\mu})
 \le\mathcal I_m(\pi_{\bar\mu})-\mathcal I_m(\pi_{\widehat\mu})
       +\mathcal R(\bar\mu)-\mathcal R(\widehat\mu)
       +\frac{m[q_\beta(\bar\mu)-q_\beta(\widehat\mu)]}{1-\beta}
       +\delta+\frac{2\varepsilon_r}{1-\beta}.
\]
We drop the two nonpositive terms involving the learned law and use $q_\beta(\bar\mu)\le\alpha$ in the signed risk difference. This gives \eqref{eq:matched-stopped} without its last term, the bound for the stopped utility. For any policy, bounded utility gives
\[
 -\frac{u_-q_\beta(\pi)}{1-\beta}
 \le V(\pi)-U(\pi)
 \le\frac{\bar u q_\beta(\pi)}{1-\beta}.
\]
We apply the upper bound to the comparison law and the lower bound to the learned law, and use $q_\beta(\bar\mu)\le\alpha$. Combining this estimate with the bound for stopped utility yields Eq.~\eqref{eq:matched-stopped}. Only the inequality for the specified
comparison law was used, so that version follows as well.
\end{proof}

\subsection{The fixed price of failure risk}

We prove each part by rearranging the identity \eqref{eq:exact-risk-scalarization} for the near-optimal policy and a comparison policy, and for part~(iv) we also account for the regularization costs and the reward approximation.

\begin{proof}[Proof of Proposition~\ref{prop:fixed-price}]
Write $\nu=m/(1-\beta)$. By \eqref{eq:exact-risk-scalarization},
$V_m(\pi)=U(\pi)-\mathcal I_m(\pi)-\nu q_\beta(\pi)$ for every
policy, with all three terms finite.
(i) If $q_\beta(\pi)\le q_\beta(\widehat\pi)$, then
\[
 U(\pi)-\mathcal I_m(\pi)=V_m(\pi)+\nu q_\beta(\pi)
 \le V_m(\widehat\pi)+\delta+\nu q_\beta(\widehat\pi)
 =U(\widehat\pi)-\mathcal I_m(\widehat\pi)+\delta .
\]
(ii) Near-optimality gives
$U(\widehat\pi)-\mathcal I_m(\widehat\pi)-\nu q_\beta(\widehat\pi)
\ge U(\pi_0)-\mathcal I_m(\pi_0)-\nu q_\beta(\pi_0)-\delta$.
Only this inequality for the one policy $\pi_0$, together with
$\widehat\pi\in\mathcal C$, is used below.
Rearranging and using $U(\widehat\pi)\le\sup_{\mathcal C}U$ and
$\mathcal I_m(\widehat\pi)\ge0$, which holds because $m\ge u_-$, proves Eq.~\eqref{eq:price-risk-law}. For
the second form, use instead
$U(\widehat\pi)\le\bar u\,\E\sum_{t<\tau}\beta^t
=\bar u[1-q_\beta(\widehat\pi)]/(1-\beta)$ and collect the terms in
$q_\beta(\widehat\pi)$. Writing
$mq_\beta(\pi_0)=(m+\bar u)q_\beta(\pi_0)-\bar uq_\beta(\pi_0)$
gives the stated form.
(iii) Put $\nu_i=m_i/(1-\beta)$ and write $q_i,U_i$ for the risk
and stopped utility of $\pi_i$. The two near-optimality conditions are
\begin{align*}
 U_2-\mathcal I_{m_2}(\pi_2)-\nu_2q_2
 &\ge U_1-\mathcal I_{m_2}(\pi_1)-\nu_2q_1-\delta_2,\\
 U_1-\mathcal I_{m_1}(\pi_1)-\nu_1q_1
 &\ge U_2-\mathcal I_{m_1}(\pi_2)-\nu_1q_2-\delta_1 .
\end{align*}
Adding them, we cancel the utilities and obtain, with
$\Delta(\pi)=\mathcal I_{m_2}(\pi)-\mathcal I_{m_1}(\pi)$,
$(\nu_2-\nu_1)(q_1-q_2)\ge\Delta(\pi_2)-\Delta(\pi_1)
-\delta_1-\delta_2\ge-\Delta(\pi_1)-\delta_1-\delta_2$, because $\Delta\ge0$, as the summand $\min\{B(z_{t+1}),u+m\}$ of Eq.~\eqref{eq:interior-loss} is nondecreasing in $m$. Division by
$\nu_2-\nu_1=(m_2-m_1)/(1-\beta)$ proves Eq.~\eqref{eq:price-monotone}.
For Eq.~\eqref{eq:price-gap-at-stake}, $r_{m_2}\le r_{m_1}$ pointwise, so $V_{m_2}\le V_{m_1}$ on $\mathcal C$ and
$\sup_{\mathcal C}V_{m_2}\le\sup_{\mathcal C}V_{m_1}\le V_{m_1}(\pi_1)+\delta_1$, while Eq.~\eqref{eq:exact-risk-scalarization} gives
$V_{m_1}(\pi_1)=V_{m_2}(\pi_1)+(\nu_2-\nu_1)q_1+\Delta(\pi_1)$. The bounds on the difference quotient hold term by term, since
$[\min\{B,u+m_2\}-\min\{B,u+m_1\}]/(m_2-m_1)$ equals one where $B\ge u+m_2$, vanishes where $B\le u+m_1$ and lies between zero and one otherwise.
(iv) Since rewards and action costs are bounded, $\widetilde\Jc_m$ is finite
on its nonempty domain $\mathcal M$ and bounded below, so all
objective comparisons concern finite real numbers. The infimum
hypothesis implies $\widetilde\Jc_m(\widehat\mu)
\le\widetilde\Jc_m(\bar\mu)+\delta$ for every $\bar\mu\in\mathcal M$;
writing it out, discarding $\mathcal R(\widehat\mu)\ge0$ and using the
reward approximation \eqref{eq:direct-certificate-step0} twice gives
\[
 V_m(\widehat\mu)\ge V_m(\bar\mu)
       -\mathcal R(\bar\mu)-\delta
       -\frac{2\varepsilon_r}{1-\beta}
 =V_m(\bar\mu)-\delta'.
\]
This is the inequality from which (ii) was derived, for
$\widehat\pi=\pi_{\widehat\mu}$, $\pi_0=\pi_{\bar\mu}$ and $\delta'$
in place of $\delta$, and $\widehat\pi$ lies in
$\mathcal C=\{\pi_\mu:\mu\in\mathcal M\}$. Repeating the two
derivations of (ii), we obtain both bounds, and $p_T\le\beta^{-T}q_\beta$
gives the finite-horizon consequence.
The bound \eqref{eq:price-supports-frontier} on the frontier follows in the same way. For $\pi\in\mathcal C$ with $q_\beta(\pi)\le\alpha$, near-optimality and
\eqref{eq:exact-risk-scalarization} give
$U(\pi)-\mathcal I_m(\pi)-\nu q_\beta(\pi)\le U(\widehat\pi)-\mathcal I_m(\widehat\pi)-\nu\widehat q+\delta$,
hence $U(\pi)-\mathcal I_m(\pi)\le U(\widehat\pi)-\mathcal I_m(\widehat\pi)+\delta+\nu(\alpha-\widehat q)$,
because $\nu\ge0$. Taking the supremum over such $\pi$ gives the upper bound on $\mathcal F_m(\alpha)$, and
the admissibility of $\widehat\pi$ itself at $\alpha=\widehat q$ gives the lower bound on $\mathcal F_m(\widehat q)$.
For $\delta=0$ the two bounds give $\mathcal F_m(\alpha)\le \mathcal F_m(\widehat q)+\nu(\alpha-\widehat q)$ for every $\alpha$,
which is the supergradient property. No concavity of the frontier is used.
\end{proof}

The hypothesis $m\ge u_-$ ensures that the interior loss is nonnegative for the clipped barrier. With a failure-only price, utility before failure is paid in full, including at zero reserve. The identity $V_m=U-mq_\beta/(1-\beta)$ then holds for every $m\ge0$. The same proof of parts~(i)--(iii) applies with $\mathcal I_m\equiv0$, for positive prices in part~(ii). The attained geometric risk of exact failure-only optimizers is nonincreasing in the price. For an interior barrier, the term $\Delta$ in part~(iii) vanishes if pre-failure barrier values do not exceed $u+m_1$.

\subsection{Action regularization and failure risk}

We bound the relative entropy between the trajectory laws through the geometric horizon by the action relative entropy and turn this bound into the stated risk bound with Pinsker's inequality.

\begin{proof}[Proof of Proposition~\ref{prop:action-risk-bridge}]
Retain states, actions and post-action environmental draws through
$\min\{T,\tau\}$, including the final consumption draw, with a common
cemetery symbol afterward. The two trajectory laws have common initial
and environmental kernels, so their likelihood ratio is the product of
conditional action-density ratios. Statewise finite relative entropy gives
$\pi_\mu\ll\rho$ and, for its density $f$,
$\int|\log f|\,d\pi_\mu\le k_\mu(y)+2/e<\infty$ at every state.
The chain rule for relative entropy therefore gives the path divergence $D_T$ between the two stopped laws through action $T$
\[
 D_T=\sum_{t=0}^T\E_{\pi_\mu}
 [\ind_{\{\tau\ge t\}}k_\mu(y_t)]\le\beta^{-T}\mathcal A(\mu).
\]
Here $\{\tau\ge t\}$ is determined before action $t$, and a failure at $t$
is retained because that action is included.
For an independent $\mathcal T$ with $\Pprob(\mathcal T=t)=(1-\beta)\beta^t$,
let $D_{\mathrm{geom}}$ denote the relative entropy between the two joint laws of $\mathcal T$ and the path through $\min\{\mathcal T,\tau\}$. Conditioning on the common horizon law and using Tonelli's theorem for the nonnegative increments gives
\[
 D_{\mathrm{geom}}=\sum_{j\ge0}(1-\beta)\beta^jD_j
 =\sum_{t\ge0}\beta^t\E_{\pi_\mu}
 [\ind_{\{\tau\ge t\}}k_\mu(y_t)]=\mathcal A(\mu).
\]
Data processing for the events $\{\tau\le T\}$ and $\{\tau\le\mathcal T\}$
gives the two binary relative-entropy bounds, since the latter probability is
$q_\beta$. The binary Pinsker inequality, $\mathrm{kl}(p\|q)\ge2(p-q)^2$,
gives the stated risk deviations, including the boundary cases.
\end{proof}

\section{Proofs of the convergence results}
\label{app:optimization-proofs}

Along any interpolation of the units, the negative clipped value and the action
relative entropy are at most $C_m$ and $\eta C_{\mathcal A}$ away from
being convex (Appendix~\ref{app:curvature}). Adding a regularizer of Proposition~\ref{prop:generators} contributes the curvature $\kappa\lambda_{\Hc}$ along generalized geodesics (Appendix~\ref{app:admissible}).
When $\Lambda_m>0$ the objective is $\Lambda_m$-convex along those
curves and the theory of \citet{ags2005} gives the gradient flow, its
uniqueness and the rates (Appendix~\ref{app:flow-proof}).

\subsection{Admissible regularizers}
\label{app:admissible}

Relative entropy to the prior $e^{-W}dx$ splits into the negative
entropy and the potential $\int W\,d\mu$. Along a generalized geodesic
the first is convex because $\log\det$ is concave on symmetric positive
definite matrices, and the second is $\lambda_W$-convex because $W$ is.
The integrand term is convex by the same change of variables when
$e^{s}G(e^{-s})$ is convex and nonincreasing. Convexity along transport
interpolations is thus different from ordinary convexity in a density.

\begin{proof}[Proof of Proposition~\ref{prop:generators}]
At $\omega=0$ we take $\Hc_{0,G}=\KL(\cdot\|\Gamma)$ on its full domain. For $\omega>0$, Jensen's inequality under the probability $\Gamma$ gives $\int G(\varrho)\,d\Gamma\ge G(\int\varrho\,d\Gamma)=G(1)$,
so the bracket in Eq.~\eqref{eq:general-divergence} is nonnegative, and
together with $\KL\ge0$ this gives $\Hc_{\omega,G}\ge\KL(\cdot\|\Gamma)\ge0$ and
$\Hc_{\omega,G}(\Gamma)=0$.

A convex $G\ge0$ with $G(0)=0$ is
continuous on $[0,\infty)$. Both $v\log v-v+1$ and $G$ are convex,
continuous, nonnegative and superlinear, so their relative integrals,
extended by $+\infty$ on measures that are not absolutely continuous
with respect to $\Gamma$, are narrowly lower semicontinuous
\citep[Lemma~9.4.3]{ags2005}, and so is their sum.

Strong convexity gives $W(x)\ge c_0+\lambda_W|x|^2/4$ for a constant $c_0$,
hence Gaussian upper tails for $\Gamma$, finite polynomial moments of all orders,
$\int W^2d\Gamma<\infty$, and the integrability of $e^{\epsilon W}e^{-W}$
for $\epsilon\in(0,1)$. The duality formula for relative integral
functionals \citep[Lemma~9.4.4]{ags2005}, for the integrand
$v\log v-v+1$ with conjugate $e^{y}-1$, applied to the bounded
continuous functions $\epsilon\min\{W,n\}$ and followed by monotone
convergence, gives
$\epsilon\int W\,d\mu\le\KL(\mu\|\Gamma)+\int e^{\epsilon W}d\Gamma-1$.
Since $W$ is bounded below, $\int|W|\,d\mu<\infty$ whenever
$\KL(\mu\|\Gamma)<\infty$, and the quadratic lower bound then gives
$\int|x|^2d\mu<\infty$, so every law of finite relative entropy to
$\Gamma$, in particular every law of finite $\Hc_{\omega,G}$, lies in
$\Ptwo$.

Let
$\Xi=p\,dx$ be a base of finite objective, and let $\mu_0,\mu_1$ be laws of
finite $\Hc_{\omega,G}$. All three are absolutely continuous with
finite second moments. Let $\mathbf t_i$ be the optimal transport maps from
$\Xi$ to $\mu_i$, put $\mathbf t_\vartheta=(1-\vartheta)\mathbf t_0+\vartheta \mathbf t_1$,
$\mu_\vartheta=(\mathbf t_\vartheta)_\#\Xi$, and
$W_\gamma^2=\int|\mathbf t_1-\mathbf t_0|^2d\Xi$, the quantity of Sect.~\ref{sec:flow} for the plan $\gamma=(\mathrm{id},\mathbf t_0,\mathbf t_1)_\#\Xi$. The resulting curve is a generalized geodesic. For an
absolutely continuous base the plan with two optimal marginals is
unique, so the curve is unique
\citep[Remark~9.2.3]{ags2005}. The maps are essentially injective,
their approximate differentials are diagonalizable with nonnegative
eigenvalues and, since the transport cost is quadratic, symmetric,
and $\det \nabla \mathbf t_\vartheta>0$ almost everywhere on $\{p>0\}$
\citep[Lemma~5.5.3, Theorems~6.2.4, 6.2.7, Proposition~6.2.12 and the
proof of Proposition~9.3.9]{ags2005}, so the change-of-variables formula
$p_\vartheta(\mathbf t_\vartheta)\det \nabla \mathbf t_\vartheta=p$ holds for the density
$p_\vartheta$ of $\mu_\vartheta$. On $\{p>0\}$ therefore
$\log(d\mu_\vartheta/d\Gamma)(\mathbf t_\vartheta(x))
=\log p(x)+W(\mathbf t_\vartheta(x))-\log\det \nabla \mathbf t_\vartheta(x)$, and
integrating against $\Xi$ gives
\[
 \KL(\mu_\vartheta\|\Gamma)
 =\int p\log p\,dx-\int\log\det \nabla \mathbf t_\vartheta\,d\Xi
  +\int W(\mathbf t_\vartheta)\,d\Xi ,
\]
whose first term does not depend on $\vartheta$. The decomposition is legitimate, since the negative part of Lebesgue
entropy is integrable under finite second moments, and
$\int|W|\,d\mu$ and the positive part of relative entropy at the
endpoints are controlled as in the previous paragraph. Concavity of
$\log\det$ on symmetric positive definite matrices and the
$\lambda_W$-convexity of $W$ along the segment
$\vartheta\mapsto \mathbf t_\vartheta(x)$ give
\[
 \KL(\mu_\vartheta\|\Gamma)
 \le(1-\vartheta)\KL(\mu_0\|\Gamma)+\vartheta\KL(\mu_1\|\Gamma)
       -\frac{\lambda_W}{2}\vartheta(1-\vartheta)W_\gamma^2 ,
\]
with finite intermediate values.

If $\omega=0$, the relative-entropy inequality proves the convexity claim on its full domain. For $\omega>0$, put $\Psi(s)=e^{s}G(e^{-s})$ along the same curve
and, on $\{p>0\}$,
$s_\vartheta(x)=-\log p(x)-W(\mathbf t_\vartheta(x))+\log\det \nabla \mathbf t_\vartheta(x)$.
Then $e^{-s_\vartheta}$ is the density of $\mu_\vartheta$ with respect
to $\Gamma$ evaluated at $\mathbf t_\vartheta(x)$, and the change of variables
$y=\mathbf t_\vartheta(x)$ gives
\[
 \int G\Bigl(\frac{d\mu_\vartheta}{d\Gamma}\Bigr)d\Gamma
 =\int_{\{p>0\}}p(x)\,\Psi(s_\vartheta(x))\,dx ,
\]
where the set on which the density vanishes drops out because $G(0)=0$.
The map $\vartheta\mapsto s_\vartheta(x)$ is concave, by the convexity
of $W$ along the segment and the concavity of $\log\det$, and $\Psi$
is convex and nonincreasing by Eq.~\eqref{eq:mccann}. Hence
$\vartheta\mapsto\Psi(s_\vartheta(x))$ is convex. This is the argument
of \citet[Theorem~9.4.12]{ags2005}, where Eq.~\eqref{eq:mccann} is stated
for positive $s$ \citep[Remark~9.3.10]{ags2005} and used for all real
$s$. Integrating the pointwise inequality gives convexity of the
integrand term and finiteness of the intermediate integrals, and adding
$\omega\ge0$ times the resulting inequality to the relative-entropy inequality gives
$\lambda_W$-convexity of $\Hc_{\omega,G}$ along generalized geodesics.

For $G(v)=v^r/(r-1)$,
$r>1$, one has $\Psi(s)=e^{-(r-1)s}/(r-1)$, which is convex and
decreasing. A nonnegative combination $G=\sum_ic_iG_i$ with some
$c_j>0$ inherits convexity, $G(0)=0$, superlinearity and the convexity and monotonicity of $s\mapsto e^{s}G(e^{-s})$ in Eq.~\eqref{eq:mccann}, the last because
$e^{s}G(e^{-s})=\sum_ic_ie^{s}G_i(e^{-s})$.

It remains to treat the potential. For $P\ge0$ continuous and $\lambda_P$-convex, $\Hc_P$ is proper, it is narrowly lower semicontinuous as the increasing limit of the integrals of the bounded continuous functions $\min\{P,n\}$, and it is $\lambda_P$-convex along every coupling interpolation of the units \citep[Proposition~9.3.2]{ags2005}, in particular along generalized geodesics. It is finite on every finitely supported law, and for quadratic weight decay, $P(x)=|x|^2/2$, one has $\lambda_P=1$.
\end{proof}

\subsection{Policy-kernel derivatives and the curvature constants}
\label{app:curvature}

We first state the kernel bounds. Take two laws $\mu_0,\mu_1\in\Ptwo$ and a
coupling $\gamma$ of them, let $\mu_\vartheta$ be the law of
$(1-\vartheta)x_0+\vartheta x_1$ under $\gamma$ for $\vartheta\in[0,1]$, and
put $W_\gamma^2=\int|x_1-x_0|^2\,d\gamma$ as in Sect.~\ref{sec:flow}. Dots denote derivatives in
$\vartheta$. In the proofs below we use only the following bounds, uniform over
states, couplings and $\vartheta$, with integral Taylor remainders uniform over states and $\vartheta$ for each coupling, namely that the policy kernels $\pi_{\mu_\vartheta}$ are twice
differentiable in total variation with
$\sup_y\|\dot\pi_{\mu_\vartheta}\|_{\rm TV}\le L_{\pi,1}W_\gamma$ and
$\sup_y\|\ddot\pi_{\mu_\vartheta}\|_{\rm TV}\le L_{\pi,2}W_\gamma^2$, and, if
$\eta>0$, the statewise relative entropy $k_\vartheta$ of
$\pi_{\mu_\vartheta}$ to $\rho$ satisfies $0\le k_\vartheta\le K_0$,
$|\dot k_\vartheta|\le K_1W_\gamma$ and $|\ddot k_\vartheta|\le K_2W_\gamma^2$. The
softmax class satisfies them under Assumption~\ref{ass:flow}(a) with
$L_{\pi,1}=L_1$, $L_{\pi,2}=2(L_2+L_1^2)$, $K_0=2L_0$, $K_1=L_0L_1$ and
$K_2=L_0L_2+(1+2L_0)L_1^2$, as shown after the proposition. The same
proofs cover any other policy class for which these bounds have been
verified.

Write $d_\beta=1-\beta$ and let $\mathsf P_\vartheta$ be the augmented transition
averaged under $\pi_{\mu_\vartheta}$. The fixed augmented transition $\mathsf P(dy'\mid y,a)$ contracts signed variation when applied to a signed action measure. Hence
$\|\dot{\mathsf P}_\vartheta\|\le L_{\pi,1}W_\gamma$ and
$\|\ddot{\mathsf P}_\vartheta\|\le L_{\pi,2}W_\gamma^2$, even for Borel dynamics and
unobserved stochastic consumption. Use the conditional reward
$\bar r_m(y,a)=\int r_m(y,a,z')\,\mathsf P(dy'\mid y,a)$.
The reward and the augmented transition are averaged over the same post-action draw.

\begin{proposition}[Value and action-law curvature]
\label{prop:flow-curvature}
Under the kernel bounds above, $-V_m$ is
semiconvex along every coupling interpolation. If the action relative-entropy bounds
are also imposed, $\mathcal A$ is semiconvex. Their deficits are
\begin{align}
 C_m&=\frac{m+\bar u}{2}
 \left(\frac{L_{\pi,2}}{d_\beta^2}+\frac{2\beta L_{\pi,1}^2}{d_\beta^3}\right),
 \nonumber\\
 C_{\mathcal A}&=\frac{K_2}{d_\beta}
 +\frac{\beta L_{\pi,2}K_0+2\beta L_{\pi,1}K_1}{d_\beta^2}
 +\frac{2\beta^2L_{\pi,1}^2K_0}{d_\beta^3}.
 \label{eq:kernel-action-curvature}
\end{align}
The corresponding functionals are $W_2$-Lipschitz, with constants
$(m+\bar u)L_{\pi,1}/(2d_\beta^2)$ and
$K_1/d_\beta+\beta L_{\pi,1}K_0/d_\beta^2$, respectively.
The action-cost conclusions are needed only when that cost is present.
\end{proposition}
\begin{proof}
Center the reward at $(\bar u-m)/2$, a constant whose value does not depend on
the policy. Put $M_0=(m+\bar u)/2$, let $b_\vartheta$ be the conditional
policy average of the centered reward, and set
$w_\vartheta=R_\vartheta b_\vartheta$ with
$R_\vartheta=(I-\beta\mathsf P_\vartheta)^{-1}$, $\|R_\vartheta\|\le d_\beta^{-1}$.
Variation-norm differentiation and the resolvent identity give
\[
 \dot w_\vartheta=R_\vartheta(\dot b_\vartheta+\beta\dot{\mathsf P}_\vartheta w_\vartheta),
 \qquad
 \ddot w_\vartheta=R_\vartheta(\ddot b_\vartheta+
 \beta\ddot{\mathsf P}_\vartheta w_\vartheta+2\beta\dot{\mathsf P}_\vartheta\dot w_\vartheta).
\]
Since $|b_\vartheta|\le M_0$ after centering, $\|w_\vartheta\|\le M_0/d_\beta$.
The kernel bounds give $\|\dot b_\vartheta\|\le M_0L_{\pi,1}W_\gamma$ and
$\|\ddot b_\vartheta\|\le M_0L_{\pi,2}W_\gamma^2$, so the first identity yields
$\|\dot w_\vartheta\|\le d_\beta^{-1}(M_0L_{\pi,1}W_\gamma+\beta L_{\pi,1}W_\gamma\,M_0/d_\beta)=M_0L_{\pi,1}W_\gamma/d_\beta^2$,
and the second gives
\[
 \|\ddot w_\vartheta\|\le\frac1{d_\beta}\Bigl[M_0L_{\pi,2}W_\gamma^2+\beta L_{\pi,2}W_\gamma^2\frac{M_0}{d_\beta}
 +2\beta L_{\pi,1}W_\gamma\frac{M_0L_{\pi,1}W_\gamma}{d_\beta^2}\Bigr]
 =M_0\Bigl[\frac{L_{\pi,2}}{d_\beta^2}+\frac{2\beta L_{\pi,1}^2}{d_\beta^3}\Bigr]W_\gamma^2,
\]
using $d_\beta+\beta=1$. The right-hand side is $C_mW_\gamma^2$.
For the action functional use the immediate state cost
$h(y)k_\vartheta(y)$, where $h(y)=\ind_{\{\min_k z_k\ge0\}}$.
This includes the failing action and excludes later ones.
Then $\|w_\vartheta\|\le K_0/d_\beta$,
$\|\dot w_\vartheta\|\le d_\beta^{-1}(K_1W_\gamma+\beta L_{\pi,1}W_\gamma\,K_0/d_\beta)$, and the
second identity gives
\[
 \|\ddot w_\vartheta\|\le\frac1{d_\beta}\Bigl[K_2W_\gamma^2+\beta L_{\pi,2}W_\gamma^2\frac{K_0}{d_\beta}
 +2\beta L_{\pi,1}W_\gamma\Bigl(\frac{K_1W_\gamma}{d_\beta}+\frac{\beta L_{\pi,1}K_0W_\gamma}{d_\beta^2}\Bigr)\Bigr]
 =C_{\mathcal A}W_\gamma^2,
\]
which is Eq.~\eqref{eq:kernel-action-curvature}.
Integrate against the fixed initial law. Lipschitz continuity follows from the first-derivative bounds along an optimal coupling. A function $g$ on $[0,1]$ whose derivative is Lipschitz with constant $L$ satisfies $g(\vartheta)\le(1-\vartheta)g(0)+\vartheta g(1)+\frac L2\vartheta(1-\vartheta)$, since $g(\vartheta)-(1-\vartheta)g(0)-\vartheta g(1)-\frac L2\vartheta(1-\vartheta)$ is convex and vanishes at both ends; applying this to $g(\vartheta)=\phi(\mu_\vartheta)$ with the second-derivative bounds gives, for $(\phi,C)=(-V_m,C_m)$ or $(\mathcal A,C_{\mathcal A})$,
\[
 \phi(\mu_\vartheta)\le(1-\vartheta)\phi(\mu_0)+\vartheta\phi(\mu_1)
 +\tfrac C2\vartheta(1-\vartheta)W_\gamma^2.
\]
Uniform remainders justify differentiation of the bounded operator
integrals. No derivative of the clipping or of the simulator is required.
\end{proof}

We now verify the bounds for softmax. Along a coupling the bounded feature
derivatives give $|\dot f|\le L_1W_\gamma$ and $|\ddot f|\le L_2W_\gamma^2$.
Bounded differentiation of the action normalizer gives
\[
 \dot\pi=\pi(\dot f-\E_\pi\dot f),\qquad
 \ddot\pi=\pi[(\dot f-\E_\pi\dot f)^2-\operatorname{Var}_\pi(\dot f)
 +\ddot f-\E_\pi\ddot f].
\]
Cauchy--Schwarz yields $L_{\pi,1}=L_1$ and $L_{\pi,2}=2(L_2+L_1^2)$.
For $k=\E_\pi f-\log\int e^fd\rho$,
\[
 \dot k=\operatorname{Cov}_\pi(f,\dot f),\quad
 \ddot k=\operatorname{Var}_\pi(\dot f)
 +\operatorname{Cov}_\pi(f,\ddot f)
 +\E_\pi[(f-\E_\pi f)(\dot f-\E_\pi\dot f)^2].
\]
Thus $K_0=2L_0$, $K_1=L_0L_1$, and
$K_2=L_0L_2+(1+2L_0)L_1^2$. The bounded derivatives provide domination. We obtain the uniform integral Taylor
remainders from the uniform Hessian continuity in Assumption~\ref{ass:flow}(a): truncate the fixed coupling to $|x_1-x_0|\le R$, use the
uniform modulus there, and bound the complement by a constant times
$\int_{|x_1-x_0|>R}|x_1-x_0|^2d\gamma$, which tends to zero.
The bounded softmax normalizer preserves these remainders. Substitution gives Eq.~\eqref{eq:flow-curvature-constant} and
\begin{equation}\label{eq:action-curvature-constants}
 C_{\mathcal A}
 =\frac{L_0(1+3\beta)L_2}{(1-\beta)^2}
 +\left\{\frac{1+2L_0}{1-\beta}
       +\frac{2\beta L_0(3-\beta)}{(1-\beta)^3}\right\}L_1^2.
\end{equation}

\subsection{Convergence of the gradient flow}
\label{app:flow-proof}

\begin{proof}[Proof of Theorem~\ref{thm:flow}]
The value and action estimates of
Proposition~\ref{prop:flow-curvature} hold for every coupling, hence
for the generalized geodesics along which the admissible regularizer
$\Hc$ is $\lambda_{\Hc}$-convex.
Adding them gives $\Lambda_m$-convexity of $\Jc_m$ along each such curve,
with the same $W_\gamma^2$, for either sign of $\Lambda_m$.
The objective is proper, $\Wtwo$-lower semicontinuous and bounded
below by $-\bar u/(1-\beta)$, hence coercive in the sense of
\citet[(2.4.10), equivalently (11.2.1b)]{ags2005}. No compactness of
its sublevels is required \citep[p.~75]{ags2005}. By
\citet[Lemma~9.2.7]{ags2005}, $\Lambda_m$-convexity along generalized
geodesics with base $\mu^j$ makes
$\mu'\mapsto\Jc_m(\mu')+\Wtwo^2(\mu',\mu^j)/(2h)$
$(\Lambda_m+1/h)$-convex along such a curve, which is the convexity
Assumption~4.0.1 of \citet{ags2005} for every $h>0$ if
$\Lambda_m\ge0$ and for $0<h<-1/\Lambda_m$ if $\Lambda_m<0$.
\citet[Lemma~4.1.1 and Theorem~4.1.2(i)]{ags2005} then give existence
and uniqueness of each minimizing-movement step from lower
semicontinuity and the completeness of $(\Ptwo,\Wtwo)$
\citep[Proposition~7.1.5]{ags2005} alone.
From the existence and contraction theory of
\citet[Theorems~4.0.4 and~11.2.1]{ags2005}
we now obtain a unique locally absolutely continuous gradient flow for
$\ell>0$, continuous in $\Wtwo$ at its initial law of finite objective,
satisfying \eqref{eq:flow-evi-main} for every $\mu'$ of finite
objective and almost every $\ell>0$.
The objective decreases along it, with
$\frac{d}{d\ell}\Jc_m(\mu_\ell)
=-|\partial\Jc_m|^2(\mu_\ell)$ almost everywhere,
where $|\partial\Jc_m|$ is the descending metric slope.
For $\Lambda_m>0$,
$\Lambda_m$-convexity along generalized geodesics gives Assumption~2.4.5 of
\citet{ags2005} through their Lemma~9.2.7 and Remark~4.0.3. Together with lower semicontinuity and coercivity, it yields existence and uniqueness of the minimizer
$\mu_m^*$ \citep[Lemma~2.4.8]{ags2005}, and, for $\mu$ of finite
objective, the slope and growth inequalities
\citep[Lemma~2.4.13]{ags2005} give
\[
 2\Lambda_m[\Jc_m(\mu)-\Jc_m(\mu_m^*)]
 \le |\partial\Jc_m|^2(\mu),\qquad
 \Jc_m(\mu)-\Jc_m(\mu_m^*)
 \ge\frac{\Lambda_m}{2}\Wtwo^2(\mu,\mu_m^*).
\]
The first inequality and the decrease of the objective give, for
$\Delta(\ell)=\Jc_m(\mu_\ell)-\Jc_m(\mu_m^*)$,
$\Delta'(\ell)=-|\partial\Jc_m|^2(\mu_\ell)\le-2\Lambda_m\Delta(\ell)$ almost
everywhere, so $e^{2\Lambda_m\ell}\Delta(\ell)$ is nonincreasing and
$\Delta(\ell)\le\Delta(0)e^{-2\Lambda_m\ell}$, the objective rate. Inserting the second into Eq.~\eqref{eq:flow-evi-main} yields
$\frac{d}{d\ell}\Wtwo^2(\mu_\ell,\mu_m^*)
\le-2\Lambda_m\Wtwo^2(\mu_\ell,\mu_m^*)$, hence the stated
distance rate. To pass to the initial time, we use continuity and
nonincrease of the objective from its finite initial value. The
contraction between two gradient flows is the $\Lambda_m$-contractive semigroup
property of \citet[Theorem~11.2.1, (11.2.2)]{ags2005}.
\end{proof}

\subsection{Optimization error and risk along the gradient flow}

\begin{proof}[Proof of Corollary~\ref{thm:combined}]
By \eqref{eq:flow-energy-rate},
$\Jc_m(\mu_\ell)\le\inf_{\mathcal M}\Jc_m+E_0e^{-2\Lambda_m\ell}$,
which is Eq.~\eqref{eq:risk-inexact-optimum} for the exact objective with
$\delta=E_0e^{-2\Lambda_m\ell}$. Apply
Proposition~\ref{prop:fixed-price}(iv), and use $\beta^Tp_T\le q_\beta$ for
the finite-horizon version.
\end{proof}

\subsection{Decreasing strength}
\label{app:decreasing-strength-proof}

\begin{proof}[Proof of Proposition~\ref{prop:decreasing-strength}]
By Proposition~\ref{prop:flow-curvature}, whose kernel bounds the softmax class satisfies under Assumption~\ref{ass:flow}(a), $-V_m$ and $\mathcal A$ are Lipschitz in $\Wtwo$. Hence $\Jc_m^{0}$ is continuous on $(\Ptwo,\Wtwo)$, and its infimum over the dense set $\mathcal M$ equals its infimum over $\Ptwo$. It is bounded below by $-\bar u/(1-\beta)$, since $V_m\le\bar u/(1-\beta)$ and $\mathcal A\ge0$.

For the regularizers of Proposition~\ref{prop:generators} the set $\mathcal M$ is dense. A law with compact support and a bounded Lebesgue density has a bounded density ratio $\varrho$ to the prior, because $W$ is continuous, so its relative entropy and $\int G(\varrho)\,d\Gamma$ are finite, $G$ being convex and finite. Its potential $\int P\,d\mu$ is finite, because $P$ is continuous. Such laws are dense in $(\Ptwo,\Wtwo)$.

For the epi-convergence, let $\mu_j\to\mu$ in $\Wtwo$. Since $\Hc\ge0$, $\Jc_m^{\kappa_j}(\mu_j)\ge\Jc_m^{0}(\mu_j)\to\Jc_m^{0}(\mu)$, which is the first condition. For the second, take $\nu_k\in\mathcal M$ with $\Wtwo(\nu_k,\mu)\le1/k$ and indices $j_1<j_2<\cdots$ with $\kappa_j\Hc(\nu_k)\le1/k$ for $j\ge j_k$, and set $\mu_j=\nu_k$ for $j_k\le j<j_{k+1}$ and $\mu_j=\nu_1$ for $j<j_1$. Then $\mu_j\to\mu$ and $\Jc_m^{\kappa_j}(\mu_j)\le\Jc_m^{0}(\nu_k)+1/k$ for $j_k\le j<j_{k+1}$, and the right side tends to $\Jc_m^{0}(\mu)$.

For the near-minimizers, a law with $\Jc_m^{\kappa_j}<\infty$ lies in $\mathcal M$, so $\mu_j\in\mathcal M$. Fix $h>0$ and $\bar\mu\in\mathcal M$ with $\Jc_m^{0}(\bar\mu)\le\inf_{\mathcal M}\Jc_m^{0}+h$. Since $\inf\Jc_m^{\kappa_j}\le\Jc_m^{\kappa_j}(\bar\mu)$,
\[
 \inf_{\mathcal M}\Jc_m^{0}\le\Jc_m^{0}(\mu_j)
 \le\inf_{\mathcal M}\Jc_m^{0}+h+\kappa_j\Hc(\bar\mu)+\delta_j-\kappa_j\Hc(\mu_j).
\]
As $\Hc(\mu_j)\ge0$, the upper limit of $\Jc_m^{0}(\mu_j)$ is at most $\inf_{\mathcal M}\Jc_m^{0}+h$. The outer inequality gives $\kappa_j\Hc(\mu_j)\le h+\kappa_j\Hc(\bar\mu)+\delta_j$, whose upper limit is at most $h$. Both limits follow, as $h$ is arbitrary. If $\mu_{j_k}\to\mu_*$ in $\Wtwo$, continuity gives $\Jc_m^{0}(\mu_*)=\lim_k\Jc_m^{0}(\mu_{j_k})=\inf_{\Ptwo}\Jc_m^{0}$.
\end{proof}

\subsection{Improvement over the prior policy}
\label{app:prior-proof}

\begin{proof}[Proof of Proposition~\ref{prop:online-baseline}]
Alignment, or $\eta=0$, gives
$\eta\mathcal A(\Gamma)=\Hc(\Gamma)=0$, so
$\Jc_m(\Gamma)=-V_m(\Gamma)$. Theorem~\ref{thm:flow} provides
the gradient flow for every sign of $\Lambda_m$, with nonincreasing objective,
so $\Jc_m(\mu_\ell)\le\Jc_m(\Gamma)$, that is,
\begin{equation}\label{eq:online-baseline-energy-control}
 \eta\mathcal A(\mu_\ell)+\kappa\Hc(\mu_\ell)
 \le V_m(\mu_\ell)-V_m(\Gamma)\le R_\Gamma .
\end{equation}
This proves the value statement. To transfer the parameter
bound to failure risk, note that for $\mu=\varrho\,\Gamma$ one has
$\|\mu-\Gamma\|_{\rm TV}=\int|\varrho-1|d\Gamma$. By hypothesis
$\|\mu-\Gamma\|_{\rm TV}^2\le c_{\Hc}\Hc(\mu)$; Pinsker and
Cauchy--Schwarz give, respectively,
\[
 \|\mu-\Gamma\|_{\rm TV}^2\le2\KL(\mu\|\Gamma),\qquad
 \|\mu-\Gamma\|_{\rm TV}^2\le\chi^2(\mu\|\Gamma),
\]
which justify $c_{\Hc}=2$ whenever $\Hc\ge\KL(\cdot\|\Gamma)$. For
$\Hc_\omega$, adding one half of the first inequality to $\omega$
times the second gives $(1/2+\omega)\|\mu-\Gamma\|_{\rm TV}^2
\le\Hc_\omega(\mu)$, that is, $c_{\Hc}=2/(1+2\omega)$.
Set $\Phi=f_\mu-f_\Gamma$ within this proof. Bounded features imply
$\|\Phi\|_\infty^2\le L_0^2c_{\Hc}\Hc(\mu)$.

At a fixed state, the log normalizer
$\Theta(\vartheta)=\log\int e^{f_\Gamma+\vartheta\Phi}d\rho$
has $\Theta''(\vartheta)=\operatorname{Var}_{\pi_\vartheta}(\Phi)
\le\|\Phi\|_\infty^2$. Differentiation under the integral is permitted
because the exponents are bounded. Integration by parts gives
\[
 \KL(\pi_\mu(\cdot\mid y)\|\pi_\Gamma(\cdot\mid y))
 =\Theta'(1)-\Theta(1)+\Theta(0)
 =\int_0^1\vartheta\,\Theta''(\vartheta)d\vartheta
 \le\tfrac12L_0^2c_{\Hc}\Hc(\mu).
\]
Because the dynamics and the initial law are common, the relative entropy
for the full path through action $T$ is the sum of these conditional action
relative entropies. This sum is at most
$(T+1)L_0^2c_{\Hc}\Hc(\mu)/2$.
The failure event is determined by that path because its post-action environmental draw, including final
consumption, is retained. The common environmental kernels cancel. Data
processing followed by the binary Pinsker bound of the proof of
Proposition~\ref{prop:action-risk-bridge} yields
\[
 |p_T(\mu)-p_T(\Gamma)|
 \le\frac{L_0}{2}\sqrt{(T+1)c_{\Hc}\Hc(\mu)}.
\]
Insert \eqref{eq:online-baseline-energy-control} to obtain the bound in Eq.~\eqref{eq:online-baseline-slack}, and solve for $\kappa$ to meet
the tolerance. When $\eta>0$,
\eqref{eq:online-baseline-energy-control} also gives
$\mathcal A(\mu_\ell)\le R_\Gamma/\eta$. Substitution into
Proposition~\ref{prop:action-risk-bridge}, with $\rho=\pi_\Gamma$,
proves the last bound.
\end{proof}

\subsection{First variations of the value and the action cost}
\label{app:first-variations}

Up to this point we have used only the curvature estimates of Proposition~\ref{prop:flow-curvature}. The first variations themselves are needed for the particle system \eqref{eq:particle} and for the proof of Proposition~\ref{prop:finite-training}. Let $v_m^\mu$ and $w_\mu$ be the statewise values of $\bar r_m$ and
of the stopped state cost $\bar k_\mu=hk_\mu$, with $h(y)=\ind_{\{\min_kz_k\ge0\}}$,
and define
\[
 \begin{aligned}
 Q_m^\mu(y,a)&=\bar r_m(y,a)+\beta\int v_m^\mu(y')\mathsf P(dy'\mid y,a),\\
 Q_{\mathcal A}^\mu(y,a)
 &=h(y)\log\frac{d\pi_\mu}{d\rho}(a\mid y)
                  +\beta\int w_\mu(y')\mathsf P(dy'\mid y,a),\\
 \mathcal O_\mu&=(1-\beta)\sum_{t\ge0}\beta^t\mathcal L_{\pi_\mu}(y_t),
 \end{aligned}
\]
where $\mathsf P$ is the augmented transition kernel, $\mathcal L_{\pi_\mu}(y_t)$ is the law of $y_t$ under $\pi_\mu$, and the occupation
measure $\mathcal O_\mu$ is normalized to mass one. A linear
functional derivative represents the derivative of
$\phi(\mu+\epsilon(\mu'-\mu))$ at $\epsilon=0$ by integration of
$\delta\phi/\delta\mu$ against $\mu'-\mu$.

\begin{proposition}[First variations]\label{prop:first-variations}
Under Assumption~\ref{ass:flow}(a), $V_m$ and $\mathcal A$ have the
bounded linear functional derivatives
\begin{align*}
 \frac{\delta V_m}{\delta\mu}(\mu,x)
 &=\frac1{1-\beta}\int\mathcal O_\mu(dy)\int\pi_\mu(da\mid y)
       [Q_m^\mu-v_m^\mu]\psi(y,a,x),\\
 \frac{\delta\mathcal A}{\delta\mu}(\mu,x)
 &=\frac1{1-\beta}\int\mathcal O_\mu(dy)\int\pi_\mu(da\mid y)
       [Q_{\mathcal A}^\mu-w_\mu]\psi(y,a,x),
\end{align*}
and their parameter gradients replace $\psi$ by $\nabla_x\psi$.
\end{proposition}

\begin{proof}
Take the affine law path $\mu+\epsilon(\mu'-\mu)$ and put
$\Delta(y,a)=\int\psi(y,a,x)(\mu'-\mu)(dx)$, so $|\Delta|\le2L_0$.
Differentiating the value through the resolvent gives the occupation integral of
$(Q_m^\mu-v_m^\mu)\Delta$. For the action cost, the derivative of the
logarithm has zero $\pi_\mu$ expectation; hence
\[
 \dot{\bar k}_\mu(y)
 =h(y)\int\log\frac{d\pi_\mu}{d\rho}(a\mid y)
   \left[\Delta(y,a)-\int\Delta(y,a')\pi_\mu(da'\mid y)\right]
   \pi_\mu(da\mid y).
\]
Combining this with the resolvent derivative $R_\mu(\dot{\bar k}_\mu+\beta\dot{\mathsf P}_{\pi_\mu}w_\mu)$, where $R_\mu=(I-\beta\mathsf P_{\pi_\mu})^{-1}$ with $\mathsf P_{\pi_\mu}$ the augmented transition averaged under $\pi_\mu$ and $\dot{\mathsf P}_{\pi_\mu}$ its derivative along the path, gives the
occupation integral of $(Q_{\mathcal A}^\mu-w_\mu)\Delta$. Here
$|Q_m^\mu-v_m^\mu|\le(m+\bar u)/(1-\beta)$,
$|\log(d\pi_\mu/d\rho)|\le2L_0$, and $\|w_\mu\|_\infty\le2L_0/(1-\beta)$.
The absolute integrability needed for all series and signed-measure
interchanges follows from these bounds, the summability of $\beta^t$
and $|\mu'-\mu|(\R^d)\le2$. Fubini yields the displayed first variations. Dominated
differentiation using $L_1$ gives their parameter gradients. The action formula includes both the change in state visitation and the derivative of the statewise divergence.
\end{proof}

\subsection{Finite-network training under a potential}
\label{app:finite-training-proof}

\begin{proof}[Proof of Proposition~\ref{prop:finite-training}]
Write $v(\mu,x)=\nabla_x[\delta(V_m-\eta\mathcal A)/\delta\mu](\mu,x)-\kappa\nabla P(x)$,
the policy gradient of Proposition~\ref{prop:first-variations} less the
gradient of the potential, so that the units move according to $\dot x_i=v(\mu_n,x_i)$. Under
Assumption~\ref{ass:flow}(a) the policy gradient is a bounded
occupation integral of an advantage times the feature gradient. By the kernel bounds of
Appendix~\ref{app:curvature}, the policy and the occupation measure are Lipschitz in the law in $\Wtwo$ as maps into total variation, and the advantage is Lipschitz in $\Wtwo$ in the supremum norm. The feature gradient is Lipschitz in
$x$ with constant $L_2$. Hence $(x_1,\dots,x_n)\mapsto(v(\mu_n,x_i))_{i\le n}$ is Lipschitz with linear growth on
$(\R^d)^n$ and the system has a unique global solution.
The curve $\mu_n(\ell)$ is locally Lipschitz in $\Wtwo$ and solves the
continuity equation with the velocity field $v_\ell=v(\mu_n(\ell),\cdot)$. At each time the field is the gradient in $x$ of the first variation of $V_m-\eta\mathcal A-\kappa\Hc_P$. A smooth cutoff equal to one near the finitely many support points gives a compactly supported potential with the same gradient at those points. Smooth approximation of that potential puts $v_\ell$ in the Wasserstein tangent space, so it is the velocity vector field of the curve.

Fix $\ell$, write $\mu=\mu_n(\ell)$, and take any $\mu'\in\Ptwo$, an
optimal coupling $\gamma$ of $\mu$ and $\mu'$, and the interpolation
$\mu_\vartheta$ of Appendix~\ref{app:curvature}, with $W_\gamma=\Wtwo(\mu,\mu')$ since $\gamma$ is optimal.
Proposition~\ref{prop:flow-curvature} gives the semiconvexity of
$-V_m+\eta\mathcal A$ along $\vartheta$ with deficit
$C_m+\eta C_{\mathcal A}$, and the convexity of $P$ gives
$\lambda_P$-convexity of $\vartheta\mapsto\int P\,d\mu_\vartheta$, so
$\vartheta\mapsto\Jc_m(\mu_\vartheta)$ is $\Lambda_m$-convex on $[0,1]$
with second differences bounded by $\Wtwo^2(\mu,\mu')$. Its right derivative at
$\vartheta=0$ is
$-\int\langle v(\mu,x_0),x_1-x_0\rangle\,d\gamma$.
This follows from the derivative of the value along the curve in the proof of
Proposition~\ref{prop:flow-curvature}, in which the score derivative at $\vartheta=0$ is
$\int\langle\nabla_x\psi(\cdot,x_0),x_1-x_0\rangle\,d\gamma$, together with Fubini's theorem, the
first variations of Proposition~\ref{prop:first-variations} and the
differentiability of $P$. The Lipschitz gradient of $P$ gives a remainder bounded by a constant times $\vartheta^2 W_\gamma^2$, so the potential derivative is integrable under the coupling. A $\Lambda_m$-convex function of $\vartheta$
lies above its tangent line corrected by the curvature, so
\[
 \Jc_m(\mu')\ge\Jc_m(\mu)
 -\int\langle v(\mu,x_0),x_1-x_0\rangle\,d\gamma
 +\frac{\Lambda_m}2\Wtwo^2(\mu,\mu').
\]
This is the subdifferential inequality of
\citet[Definition~10.3.1]{ags2005} for the plan carried by the field
$-v(\mu,\cdot)=-v_\ell$, with the quadratic term absorbed
in its remainder. Hence $v_\ell$ belongs to minus the subdifferential of
$\Jc_m$ at $\mu_n(\ell)$ for every $\ell$, and $\mu_n(\cdot)$ is a
gradient flow in the sense of \citet[Definition~11.1.1]{ags2005}.
The objective is lower semicontinuous and $\Lambda_m$-convex along
generalized geodesics, in particular along geodesics, so by
\citet[Theorem~11.1.4]{ags2005} the gradient flow from $\mu_n(0)$ is
unique, hence equal to the gradient flow of Theorem~\ref{thm:flow}, and Eq.~\eqref{eq:finite-training} is the contraction (11.1.19) there
against the gradient flow from $\mu_0$. The rate toward $\mu_m^*$ is Eq.~\eqref{eq:flow-distance-rate} applied from $\mu_n(0)$.

For the collapse, let $\Lambda_m>0$ and apply the correspondence just proved with $n=1$. The curve of point masses $\delta_{x(\ell)}$ with $\dot x=v(\delta_x,x)$ is the gradient flow of Theorem~\ref{thm:flow} from $\delta_{x_0}$ and converges in $\Wtwo$ to the unique minimizer $\mu_m^*$. A $\Wtwo$-limit of point masses is a point mass. Hence $\mu_m^*=\delta_{x^*}$, and $x^*$ minimizes $x\mapsto\Jc_m(\delta_x)=-V_m(\delta_x)+\eta\mathcal A(\delta_x)+\kappa P(x)$, since $\mu_m^*$ minimizes $\Jc_m$ over all laws and in particular over point masses. By the last contraction of the statement, every unit then converges to $x^*$.

The value bound follows from the $\Wtwo$-Lipschitz constant of
Proposition~\ref{prop:flow-curvature}. For the geometric risk
write $q_\beta(\pi)=(1-\beta)\sum_t\beta^t\,\mathbb P_\pi(\tau\le t)$
and $\mathbb P_\pi(\tau\le t)=\E_\pi b(y_t,a_t)$, which holds since a negative
reserve stays negative. Here the conditional failure cost is
$b(y,a)=1$ if $\min_kz_k<0$ and otherwise the
$\mathsf P(\cdot\mid y,a)$-probability that the next reserve is
negative, a cost in $[0,1]$ like $\bar r_m$. Centering it at $1/2$ and repeating the resolvent
computation in the proof of that proposition with $M_0=1/2$ gives
$|\dot w_\vartheta|\le L_1W_\gamma/(2d_\beta^2)$, hence the stated constant
after multiplication by $1-\beta$.
\end{proof}

\section{Proof of the population-to-network transfer}
\label{app:deployment-proofs}

When $n$ units are sampled, the score is perturbed by an average of $n$ bounded
centered terms. The relative entropy between the trajectory laws of
the population policy and of the sampled network, up to an independent
geometric time, is then of order $1/n$, and we turn this estimate into
a total-variation bound on value and risk with Pinsker's inequality.

\begin{proof}[Proof of Theorem~\ref{thm:finite-deployment}]
Write $\mathbb E_X$ for expectation over the sampled units and
$\Phi_n(y,a)=f_n(y,a)-f_\mu(y,a)$. At every fixed state $y$, the
normalizing constants in the two action kernels give
\begin{equation}\label{eq:compact-finite-kl}
 K_n(y):=\KL(\pi_\mu(\cdot\mid y)\|\pi_n(\cdot\mid y))
 =\log\int e^{\Phi_n}d\pi_\mu-\int\Phi_n\,d\pi_\mu.
\end{equation}
For fixed $(y,a)$, $\Phi_n$ averages $n$ independent centered
variables with range length $2L_0$. By independence, $\mathbb E_X\Phi_n=0$ and $\mathbb E_Xe^{\Phi_n}$ is the product of $n$ factors $\mathbb E\,e^{Z_i/n}$, where $Z_i=\psi(y,a,X_i)-f_\mu(y,a)$ is centered with range $2L_0$. The bounded-variable exponential estimate underlying Hoeffding's inequality \citep{hoeffding1963}, $\mathbb E\,e^{\lambda Z}\le e^{\lambda^2(2L_0)^2/8}$, applied with $\lambda=1/n$, gives $\mathbb E_X e^{\Phi_n}\le e^{n\cdot4L_0^2/(8n^2)}=e^{L_0^2/(2n)}$.
Since $|\Phi_n|\le2L_0$, Fubini applies to its signed integral.
The exponential is nonnegative and jointly measurable, so Tonelli
applies to its integral against the product of the parameter law and the
fixed kernel $\pi_\mu(\cdot\mid y)$. Jensen's inequality for the
logarithm in Eq.~\eqref{eq:compact-finite-kl} therefore yields
$\mathbb E_X K_n(y)\le L_0^2/(2n)$ uniformly in $y$.

Let $\mathcal T$ be independent with $\mathbb P(\mathcal T=t)=(1-\beta)\beta^t$.
Denote by $\Pprob_\mu,\Pprob_n$ the laws of the trajectory through action $\mathcal T$,
including $\mathcal T$ and the post-action environmental draws, where the
latter law is conditional on the sampled units.
Their initial, environmental and horizon laws agree. The failure
event is measurable because the final consumption draw is retained. Their action-density
ratios multiply along a finite trajectory, giving the chain rule
\begin{equation}\label{eq:compact-finite-path-kl}
 D_n:=\KL(\Pprob_\mu\|\Pprob_n)
 =\sum_{t\ge0}\beta^t\mathbb E_{\pi_\mu}K_n(y_t),
 \qquad
 \mathbb E_XD_n\le\frac{L_0^2}{2n(1-\beta)}.
\end{equation}
The one-step log density ratio is bounded by $4L_0$, so the signed
logarithms are absolutely integrable, bounded in expectation by
$4L_0/(1-\beta)$. This bound justifies taking expectations in the chain rule.
After conditioning, all $K_n$ terms are nonnegative. Tonelli's theorem, applied over time,
the population trajectory law and the independent parameter law, justifies
the exchanges in Eq.~\eqref{eq:compact-finite-path-kl}. The population trajectory law is independent of the sampled units.

If $L_0=0$, the two policies coincide. Otherwise Markov's inequality gives $D_n\le L_0^2/(2n(1-\beta)\zeta)$ with probability at least $1-\zeta$ over the sampled units, and Pinsker's inequality $\mathrm{tv}\le\sqrt{D_n/2}$ then gives
\[
 \mathrm{tv}(\Pprob_\mu,\Pprob_n)
 \le\sqrt{D_n/2}\le
 \frac{L_0}{2\sqrt{n(1-\beta)\zeta}},
\]
where $\mathrm{tv}(\Pprob,\Pprob')=\sup_E|\Pprob(E)-\Pprob'(E)|\le1$.
The event $\{\tau\le\mathcal T\}$ is determined by the trajectory through
action $\mathcal T$, and its probability is $q_\beta$. Moreover
\begin{gather*}
 (1-\beta)V_m(\pi)=\mathbb E_\pi r_m(y_{\mathcal T},a_{\mathcal T}),\qquad
 (1-\beta)V(\pi)=\mathbb E_\pi u(s_{\mathcal T},a_{\mathcal T}),\\
 (1-\beta)U(\pi)=\mathbb E_\pi[u(s_{\mathcal T},a_{\mathcal T})\ind_{\{\mathcal T<\tau\}}].
\end{gather*}
These identities follow by Fubini, since each discounted absolute sum
is bounded by the corresponding uniform reward bound divided by $1-\beta$.
The geometric event has range length one. The stopped utility at the geometric time lies in $[-u_-,\bar u]$ because this interval contains zero. Expectations of a function with range length $L$ under the two laws differ by at most
$L\,\mathrm{tv}$, proving the three geometric-horizon conclusions.
For the deterministic horizon $T$, the same chain rule has $T+1$
terms, so the expected relative entropy is at most
$(T+1)L_0^2/(2n)$. The Markov and Pinsker inequalities give Eq.~\eqref{eq:finite-direct-horizon}. For the simultaneous statement, we apply a union bound with error $\zeta/2$ for each construction event.
\end{proof}

\section*{Acknowledgements}
We acknowledge CSC -- IT Center for Science, Finland, for computational resources.

\renewcommand{\bibfont}{\linespread{0.96}\selectfont}
\bibliographystyle{abbrvnat}
\bibliography{references}

\end{document}